\documentclass{article}

\usepackage{arxiv} 

\usepackage{url,hyperref,lineno,microtype,subcaption}
\usepackage[onehalfspacing]{setspace}

\usepackage{float}  
\usepackage{amsmath} 
\usepackage{gensymb} 

\usepackage{natbib}  % \citep
\usepackage{graphicx}  % \includegraphics

\title{Embodiment Enables Non-Predictive Ways of Coping with Self-Caused Sensory Stimuli}
\renewcommand{\shorttitle}{Coping with Self-Caused Sensory Stimuli}

\author{James Garner $^{1}$ \\
    \texttt{james.garner@auckland.ac.nz} \\
    \\
    $^{1}$ School of Computer Science \\
    University of Auckland \\
    Auckland, New Zealand \\
    \And
    Matthew Egbert $^{1,2}$ \\
    \\
    \\  
    $^{2}$Te Ao M\=arama -- Centre for Fundamental Inquiry \\
    The University of Auckland \\
    Auckland, New Zealand \\
}

\begin{document} 
\maketitle

\begin{abstract}

Living systems process sensory data to facilitate adaptive behaviour.
A given sensor can be stimulated as the result of internally driven activity, or by purely external (environmental) sources.
It is clear that these inputs are processed differently - have you ever tried tickling yourself?
The canonical explanation of this difference is that when the brain sends a signal that would result in motor activity,
it uses a copy of that signal to predict the sensory consequences of the resulting motor activity.
The predicted sensory input is then subtracted from the actual sensory input,
resulting in attenuation of the stimuli.
To critically evaluate this idea,
and investigate when non-predictive solutions may be viable,
we implement a computational model of a simple embodied system with self-caused sensorimotor dynamics,
and analyse how controllers successfully accomplish tasks in this model.
We find that in these simple systems, solutions that regulate behaviour to control self-caused sensory inputs tend to emerge, rather than solutions which predict and filter out self-caused inputs.
In some cases, solutions depend on the presence of these self-caused inputs.

\end{abstract}

\section{Introduction}

The remarkable adaptive behaviour displayed by living organisms would not be possible without the capacity to respond to sensory stimuli appropriately.
The same sensors can be stimulated due to external (environmental) causes, as well as by internally driven activity.
Intuitively, it seems like responding appropriately must require distinguishing the two.
We can hear sounds in the world around us, but we can also hear our own voice when talking, and our own footsteps when walking. We can see our environment, but we also see our own bodies. Not only do we perceive both the world and the results of our own actions, but the exact same sensory stimulus can be the result of an external event, or caused by our own activity.
For example the sight of a hand being waved before our eyes could be your own hand or a friend snapping you out of a daydream.
However, we typically have no trouble telling the difference.
Indeed, the phenomenology of a self-caused stimuli can be very different from that of an externally caused one.
A great example of this is the sensation of touch, which can reduce you to helpless laughter when externally applied - but trying to tickle yourself just isn't the same! \citep{blakemore_why_2000}
Understanding exactly how these inputs are processed differently can facilitate building artificial systems as capable and flexible as living ones.

In psychology, research on the sensory attenuation of self-caused stimuli studies how these stimuli are perceived as diminished in comparison to externally caused stimuli \citep{hughes_erp_2011}. A clear example of this effect is seen in the force-matching paradigm. Here an external force is applied to a subject’s finger, after which they must use their other hand to recreate that force as precisely as possible. This takes place under two conditions. In the direct condition, the subject applies force to their finger in a manner as close as possible to pressing on their own finger. In the indirect condition, they apply the force via a mechanism, such as a lever to one side. Healthy subjects consistently apply too much force when pressing directly on their finger, indicating that the perceived force is attenuated compared to the other conditions \citep{parees_loss_2014}.
The canonical explanation of this effect is that when the brain issues a motor command, it uses a copy of that command to predict the sensory consequences of the resulting motor activity. The predicted sensory input is then subtracted from the actual sensory input, resulting in the attenuation of the stimuli \citep{klaffehn_sensory_2019}. 

The problem of ego-noise in robotics hints at why subtracting out self-produced stimuli seems like a natural thing for the brain to do. Ego-noise refers to self caused noise, including that of motors. This noise can interfere with the data collecting sensors of a robot, and the straightforward engineering solution is to cancel out the noise. The explicitly representational and predictive explanation of the sensory attenuation effect meshes well with this engineering perspective, and has informed a predictive approach to dealing with ego-noise \citep{schillaci_body_2016}.

The aim of this paper is to critically evaluate the idea that these representational, predictive and subtractive approaches are the natural solution to the problem of self-caused sensory inputs interfering with perceiving the world. To do so, we developed a computational model of a simple embodied system with non-trivial self-caused sensorimotor dynamics. Using this model, we examined under what conditions different ways of succeeding at tasks despite self-caused sensory interference might emerge.

The embodiment is a simple, simulated, two-wheeled system with a pair of light sensors. It is coupled to a controller - a continuous-time, recurrent neural network (CTRNN) - which determines its motor activity. The sensory input to this robot is a linear combination of environmental factors (a function of its position relative to a light) and a self-caused component - a function of the robot’s motor activity.

This model is designed to allow for both representationalist and non-representationalist solutions to emerge. Because the interfering dynamics are produced by a simple, smooth function, they could be fully modelled by the CTRNN controller \citep{beer_parameter_2006}. And because the interfering dynamics are summed with the actual sensor data, the problem can be solved by predicting them and subtracting them out. This explicitly representational solution would fit with the canonical explanation of sensory attenuation. However, as the interfering dynamics are a function of the system’s motor activity, and are coupled to the controller in a tight sensorimotor loop, this model embraces the situated, embodied and dynamical approach, and also allows for the emergence of other non-representationalist solutions.

Following the evolutionary robotics methodology
we explore the space of possible solutions using a genetic algorithm
\citep{harvey_evolutionary_2005}.
We then analyse the behavioural strategies of controllers tuned to successfully accomplish a task (phototaxis).
This permits us examine a range of ways embodied systems may cope with self-caused sensory stimuli, and gain insight into the conditions in which alternatives to representational and predictive solutions may emerge.

In Section \ref{sec_method} we explain the model we developed and how we optimise its parameters.
Then in Section \ref{sec_experiments} we present the results of our investigation.
Finally, we summarise our findings and their relevance in Section \ref{sec_discussion}.

\section{Model and methods}
\label{sec_method}

In this section we describe the model and optimisation method we use to investigate how embodied systems cope with self-caused sensory input.

\subsection{Model}
We use a model of a simple light-sensing robot, controlled by a neural network, where motor activity can directly stimulate the robot's light sensors.
The two wheeled robot moves about an infinite, flat plane. It has a pair of directional light sensors, and the environment contains a single unchanging light source.
The robot is controlled by a continuous-time, recurrent neural network (CTRNN).
Motor-driven interference is ipsilateral and non-saturating, and is determined by one of three different functions, which are detailed in the Experiments section.

\subsubsection{Embodiment}

The robot is circular, with two idealised wheels situated on its perimeter $\pi$ radians apart, at $-\pi/2$ and $\pi/2$ relative to its facing.
The wheels can be independently driven forwards or backwards. 
Its two light sensors are located on its perimeter at $-\pi/3$ and $\pi/3$ relative to its facing.
The environment it inhabits is defined entirely by the spatial coordinates of the single light source.
The robot's movement in its environment is described by the following set of equations:

\begin{alignat}{2}
    \dot{x} &= (m_L + m_R)\cos(\alpha)
    \\
    \dot{y} &= (m_L + m_R)\sin(\alpha)
    \\
    \dot{\alpha} &= (m_R - m_L)r\label{eq_robot_alpha}
\end{alignat}

Where $x$ and $y$ are the robot's spatial coordinates,
and $\alpha$ is the robot's facing in radians.
$m_L$ and $m_R$ are the robot's left and right motor activation respectively,
and are always in the range $[-1, 1]$.
The values of $m_L$ and $m_R$ are specified by the controller, which is described later.
$r=0.25$ is the robot's radius.
We simulate this system using Euler integration with $\Delta_t=0.01$.

Physically this describes positive motor activation turning its respective wheel forwards, and conversely for negative motor activation.
If the sum of the two motors' activation is positive, the robot as a whole moves forwards with respect to its facing, while if it is negative, the robot moves backwards.
The amount that the robot turns is also determined by the relationship between the two wheels.

The robot's two light sensors are located at the coordinates $x + cos(\alpha + \theta)r, y + sin(\alpha + \theta)r$,
where $\theta$ is the sensor's angular offset.
For the left sensor, $\theta = \pi/3$ and for the right sensor, $\theta = -\pi/3$.
The environmental stimulation of the sensors is given by: 

\begin{equation}
    s = \frac{(\boldsymbol{b} \cdot \boldsymbol{\hat{c}})^+}{1 + D^2} \boldsymbol{\epsilon}
\end{equation}

Where $\boldsymbol{b} = [cos(\alpha + \theta), sin(\alpha + \theta)]$
is the unit vector pointing in the direction the sensor is facing,
and $\boldsymbol{c}$ is the vector from the sensor to the light,
with $\boldsymbol{\hat{c}}$ denoting the vector is normalised to have a unit length.
That is $\boldsymbol{\hat{c}}=\boldsymbol{c} / |\boldsymbol{c}|$,
where $|\boldsymbol{c}|$ is the magnitude of $\boldsymbol{c}$.
The symbol $\cdot$ denotes the dot product of the two vectors,
and the superscript $^+$ indicates that any negative values are replaced with $0$.
$D$ is the Euclidean distance from the sensor to the light, and $\boldsymbol{\epsilon}=5$ is a fixed environmental intensity factor.
$s_L$ denotes the activation of the left sensor, with $\theta=\pi/3$,
while $s_R$ denotes the activation of the right sensor, $\theta=-\pi/3$.

The numerator is maximised at $1$ when the sensor is directly facing the light,
and minimised at $0$ when the sensor is facing $\pi/2$ radians (90\degree) away from the light.
The denominator is minimised at 1 when the distance from the sensor to the light is 0.
This means that the activation of the sensor grows both as the sensor faces more towards the light,
and as the sensor approaches the light
(so long as it is facing less than $\pi/2$ radians away from the light).

\subsubsection{Controller}

The controller is a continuous-time recurrent neural network (CTRNN) defined by the state equation below, following \cite{beer_toward_1996}:

\begin{equation}
    \tau_i \dot{y_i} = -y_i + \sum_{j=1}^{N} \omega_{[j \rightarrow i]} \sigma (y_j + \beta_j) + I_i
    \label{eq_ctrnn}
\end{equation}

Here $N=10$ denotes the number of neurons in the network.
$y_i$ indicates the activation of the $i$th neuron.
The parameter $\tau_i$ is its time constant, where $0 < \tau_i < 3$, 
while the parameter $\beta_i$ is its bias, where $-5 < \beta_i < 5$.
$I_i$ is any external input to the neuron.
$\sigma(x) = 1/(1 + e^x)$ is the standard logistic activation function for neural networks, and is a sigmoid function $\in [0, 1]$.
$\omega_{[j \rightarrow i]}$ is a weight determining the influence of the $j$th neuron on the $i$th neuron, where $-5 < \omega_{[j \rightarrow i]} < 5$.

Some neurons are designated output neurons, and their activation values $y$ are treated as the output of the network.
In our case, $y_{N-1}$ and $y_{N}$ provide the values $m_L$ and $m_R$ respectively.
Output is scaled to be $\in [-1, 1]$ by the function:
\begin{equation}
    o(y) = \frac{2}{1 + \exp({\frac{-y}{\sqrt{\omega_{\text{max}}}}})} - 1 
\end{equation}

Where $\omega_{\text{max}}=5$ denotes the maximum weight value $\omega$ permitted for a node in this CTRNN.

Some neurons are designated input neurons, and all their incoming interneuron weights $\omega_{[j \rightarrow i]}$ are set to 0, including the recurrent weight $\omega_{[i \rightarrow i]}$.
With the robot described above, neurons 1 and 2 are designated as input neurons, and $I_1 = \omega_I s_L$, while $I_2 = \omega_I s_R$, where $\omega_I = 5$ is a fixed input scaling weight.
These are the only neurons which receive an external input,
so $I_{3..N} = 0$ always.

\subsubsection{Motor-driven interference}

Perception necessarily involves both the system and its environment.
Nevertheless, we can consider the degree to which the activity of system or environment contribute to a given stimuli.
Let us take three very different points on this spectrum.
(1) If our robot passively sat still, while a light in the environment turned on and off, the change in the light sensors' activations would primarily be due to external causes.
(2) On the other hand, in the model described above, all changes in the light sensors' activations are the result in a change in the relationship between the light's position and the robot's position and facing.
(3) At the other end of the scale, if the robot could take an action like using an arm to shine a light source into its sensors, then the change in the sensors' activation would be due primarily to the robot's own activity.
Our perception covers this spectrum. Consider the visual experience of (1) sitting passively and watching a movie, (2) turning to look out the window briefly, then (3) scratching your nose.

In the model described so far, there is no possibility for directly self-caused stimuli like (3).
This is the kind of self-caused sensory input we are concerned with, so we extend the model with an interference function $\psi(m)$.
The various interference functions we study are described in Section \ref{sec_experiments}.
The interference function is used in a new sensory input equation:

\begin{equation}
    s' = \lambda\psi(m) + (1-{\lambda})s
\end{equation}

Where $s$ is the original light sensor activation,
$m$ is the ipsilateral motor's output,
and $\lambda$ is a scaling term controlling how much of the sensory input is due to the environment, and how much is due to the system's motor activity. Substituting for the original input neuron equations, this gives:

\begin{align}
    I_1 = \omega_Is'_L = \omega_I({\lambda}\psi(m_L) + (1-\lambda)s_L)
    \\                                              
    I_2 = \omega_Is'_R = \omega_I({\lambda}\psi(m_R) + (1-\lambda)s_R)
\end{align}

This combination of motor-driven interference with sensor activity is additive and non-saturating. That is, the interference $\psi(m)$ can never be so high that change in the environmental stimulation $s$ of the sensor does not result in a change in $s'$.
This means that if $\psi(m)$ can be predicted by the network, then this value can simply be subtracted from the input neuron's output to other nodes.
This mapping also uses the ipsilateral motor to generate interference for each sensor.
This was chosen for two reasons.
Firstly, it is physically intuitive.
Secondly, because the motor neurons have recurrent connections to the interneurons,
this means that the inputs used for $\psi(m)$ are available internally to the network, making prediction possible.

To summarise, we start with a model of a 2 wheeled robot with 2 light sensors, controlled by a CTRNN.
In this model, changes in a light sensor's activation are purely the result of the robot's position and orientation changing relative to the light.
We extend this model by adding a function which, given a motor activation value, produces an interfering output.
Instead of the input neurons of the controller directly receiving the current activation of the light sensor,
the light sensor's activation is first combined with this interference.
The parameter $\lambda$ controls the weighting given to the sensor activation vs the interference in this combined term.
For example, with $\lambda = 0.05$, instead of the light sensor's true reading $s$, the controller receives $0.95 s + 0.05 \psi(m)$.
The interference functions $\psi(m)$ are described in the Experiments section.

\subsection{Methods}

Parameters for the CTRNN controller were evolved using a tournament based genetic algorithm (GA) based on the microbial GA \citep{harvey_microbial_2009}.
The GA operates on a population, which consists of a number of solutions specifying the parameters for the CTRNN.
In a tournament, two randomly chosen solutions from the population are evaluated independently.
Their fitness is compared, and then in the reproduction step the lower scoring solution is removed from the population and replaced by a mutated copy of the higher scoring solution.
Our microbial GA differs from the classic presentation in that it ensures that each member of the population participates in exactly one tournament before the reproduction step is performed for the entire population.
This allows generations of the population to easily be counted.

The following parameters were evolved for each node $i$ in the CTRNN: the time factor $\tau_i$, the bias $\beta_i$, and a weight vector $\omega_{ji}$, giving the incoming weight given to connections to this node from all other nodes in the network.

In all cases the system was evolved to perform phototaxis using the following fitness function:

\begin{equation}
    \frac  { \sum_{t=0}^{T} d(x_t, y_t)^2t }  { \sum_{t=0}^{T} t }
    \label{eq_fitness_func}
\end{equation}

Where $t$ is the time at the current integration step, $T$ is the trial duration, and  $d(x, y)$ is the euclidean distance from the point $(x, y)$ to the light.
The squared distance is used rather then the actual distance here solely for computational efficiency.
Multiplying the distance by the current time means that minimising distance later in the trial is more important to the fitness score than doing so earlier is.
The final distance is the most important, while the original distance from the light at $t=0$ is completely disregarded. However, improvement at any time is always relevant: $t=99$ is almost as important as $t=100$.

In each trial, the robot begins at the origin.
Each generation, four light coordinates are stochastically generated.
The first coordinate is chosen uniformly at random to lie on a circle of radius 3 centred on the origin.
The other three coordinates lie on the same circle and form a square with the first.
Each solution in the population has its fitness score calculated for each of the four light coordinates.
These scores are combined before comparison in the tournament.
This means that a given solution's score may go up or down from generation to generation, as it may perform better or worse on that generation's set of light coordinates.
This helps prevent the GA becoming stuck in a local optima.

A population of 50 individuals was used. The trial duration was chosen to allow enough time for robust phototaxis to be selected for, either 10 or 20 time units depending on the interference function. The GA was allowed to run for a sufficient number of generations for fitness gains to plateau and for the population of solutions to converge.

\section{Experiments}
\label{sec_experiments}

To investigate how embodied systems cope with motor-driven interference,
we began by using the GA to find parameters that would allow a CTRNN controller to perform phototaxis in the basic model with $\lambda=0$ (i.e. with no motor interference).
The population of controllers that were the product of this GA run are taken as the \textit{ancestral population} for the subsequently evolved populations in Experiments 2-4.
That is, parameters for these populations were evolved starting from this ancestral population, rather than starting from a new, random population.
We chose to use an ancestral population, rather than evolving subsequent populations from scratch, in order to allow for direct comparison between the behaviour of the systems optimised with and without the presence of motor-driven interference.

In addition to Experiment 1 with the basic version of the model where $\lambda=0$ (and therefore $s' = s$), we consider three further versions of the model in Experiments 2-4, each corresponding to a different interference function $\psi(m)$.
We use $\lambda=0.5$ with each of these three functions.

\subsection{Experiment 1: Phototaxis Without Interference}

\begin{figure}[t]
\begin{center}
\includegraphics[width=0.8\textwidth]{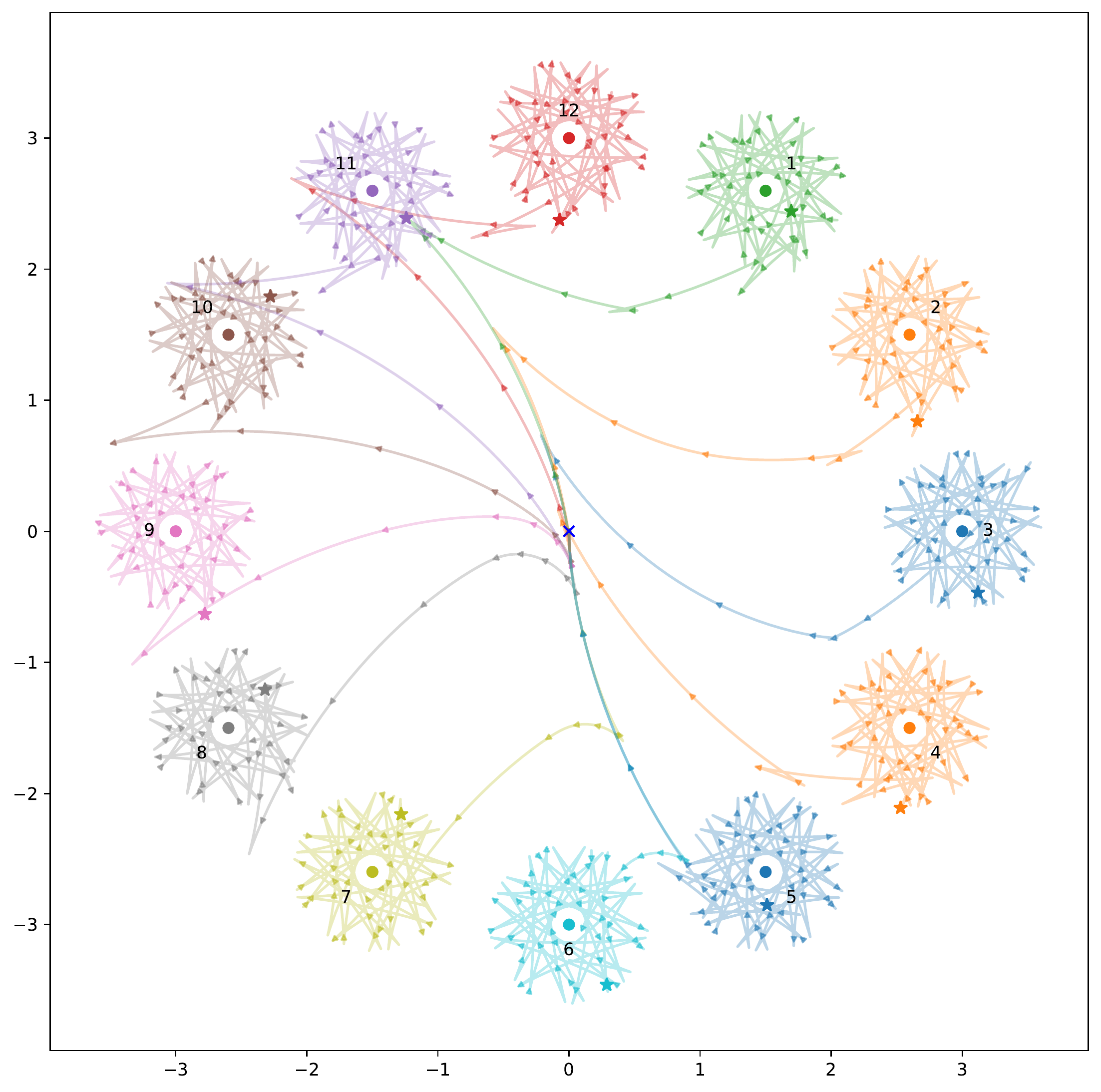}
\end{center}
\caption{
    Spatial trajectories for the best individual from the ancestral population for 12 different light coordinates.
    The robot always begins at the origin, facing towards positive $y$.
    Stars mark the final position reached during the trial duration used during evolution.
    The coloured circles show the light position for the correspondingly coloured trajectory.
    The triangles along the trajectories point in the direction the robot is facing.
}
\label{fig_ancestor_5k_trajectories}
\end{figure}

A highly fit population of controllers was evolved to perform phototaxis in the basic model, with no motor-driven interference.
Evolution of this population began from a population of solutions with uniformly random interneuron weight
and time constant
values, and with centre-crossing biases
\citep{mathayomchan_center_2002}.
A trial duration of 10 time units was used.
After evolution, genomes for this population are highly convergent, indicating that the population has become dominated by a single solution.
Examining the fittest member of this population, we found that the controller reliably brought the robot close to the light across a collection of light coordinates representative of those used during evolution (Figure \ref{fig_ancestor_5k_trajectories}).
The robot's behaviour results in it remaining close to the light even over time periods orders of magnitude longer than the trial duration used during evolution.
This indicates that the solution produces a long term, stable relationship with the environmental stimuli.

The ancestral solution's behaviour is well preserved in the descendent populations evolved to handle the various interference functions studied.
Understanding how this solution works is helpful for understanding how the descendent solutions handle motor-driven sensory interference.

The ancestral solution's behaviour can be divided into 2 phases:

\begin{enumerate}
    \item[A)] The \textit{approach} phase, where the robot makes its way close to the light. This phase has to account for the light starting at an unknown point relative to the robot.

    \item[B)] The \textit{orbit} phase, where the robot's long-term periodic activity maintains a close position to the light.
\end{enumerate}

Note that this two phase description does not imply switching between two different sets of internal rules.
These phases are driven by the ongoing relationship between the robot and its environment, and are better thought of in dynamical systems terms as a transient and a periodic attractor.

The orbit phase (Phase B) is simpler to explain, so we will begin with it. Here we can approximate the robot's behaviour with a simple program:

\begin{figure}[t]
\begin{center}
\includegraphics[width=\textwidth]{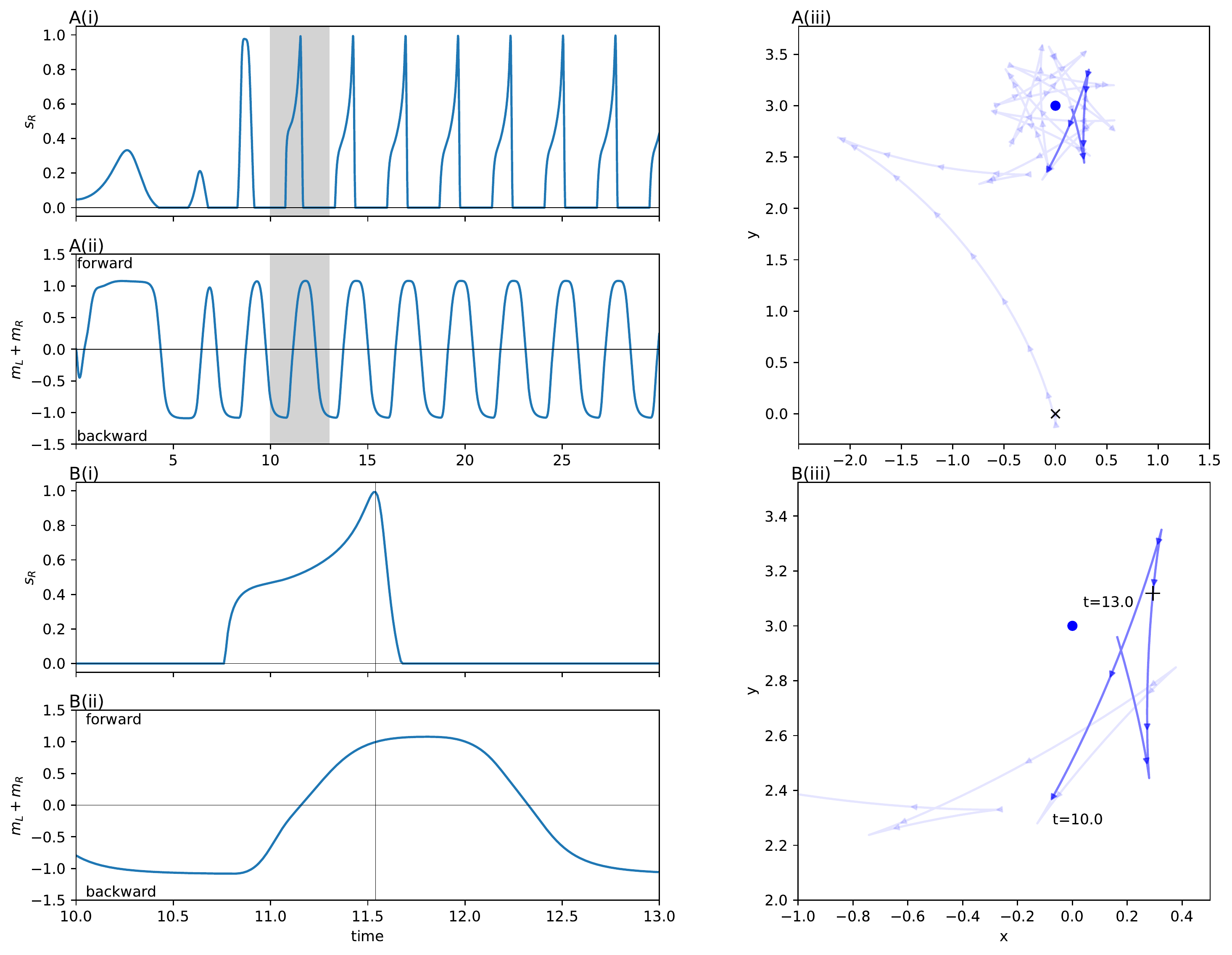}
\end{center}
\caption{
    A plots show the ancestral solution when the light is at coordinates (0, 3) - position 12 in Figure \ref{fig_ancestor_5k_trajectories}.
    The highlighted sections of these figures mark the time period 10-13, which is shown in more detail in B.
    The vertical line in B(i) and B(ii) marks the peak of right sensor activation,
    which corresponds to the + in B(iii).
    The activity shown in B corresponds to the Phase B program (see main text).
    Before $t=11$, the robot drives backwards, passing the light on its right side.
    As the right sensor is stimulated, the robot changes direction, driving forwards.
    After the right sensor stimulation peaks and dies down, the robot changes direction again, reversing towards the light.
    A show how the process repeats.
}
\label{fig_ancestor_detail_right}
\end{figure}

\begin{enumerate}
    \item Approach the light while driving backwards, such that you will pass the light with the light on your right hand side.

    \item When the light abruptly enters your field of vision, it causes a spike in your right sensor. Quickly respond by switching to driving forwards instead, turning gently to the left.

    \item After driving forward has brought the light behind you and out of the sensor's field, go to 1.
\end{enumerate}

We observe this behaviour across all the light coordinates we examined.
Figure \ref{fig_ancestor_detail_right} and the corresponding caption explains how this behaviour applies to the trajectory for a specific light coordinate,
showing how the simple program described above matches its behaviour.
The left sensor is completely uninvolved in this process. In fact for some initial light positions, namely when the robot begins with the light on its right, the left sensor is also completely uninvolved in the approach phase. That is, if the left sensor is completely deactivated throughout certain trials, the trajectory is completely identical to if it were active.

\begin{figure}[t]
\begin{center}
\includegraphics[width=\textwidth]{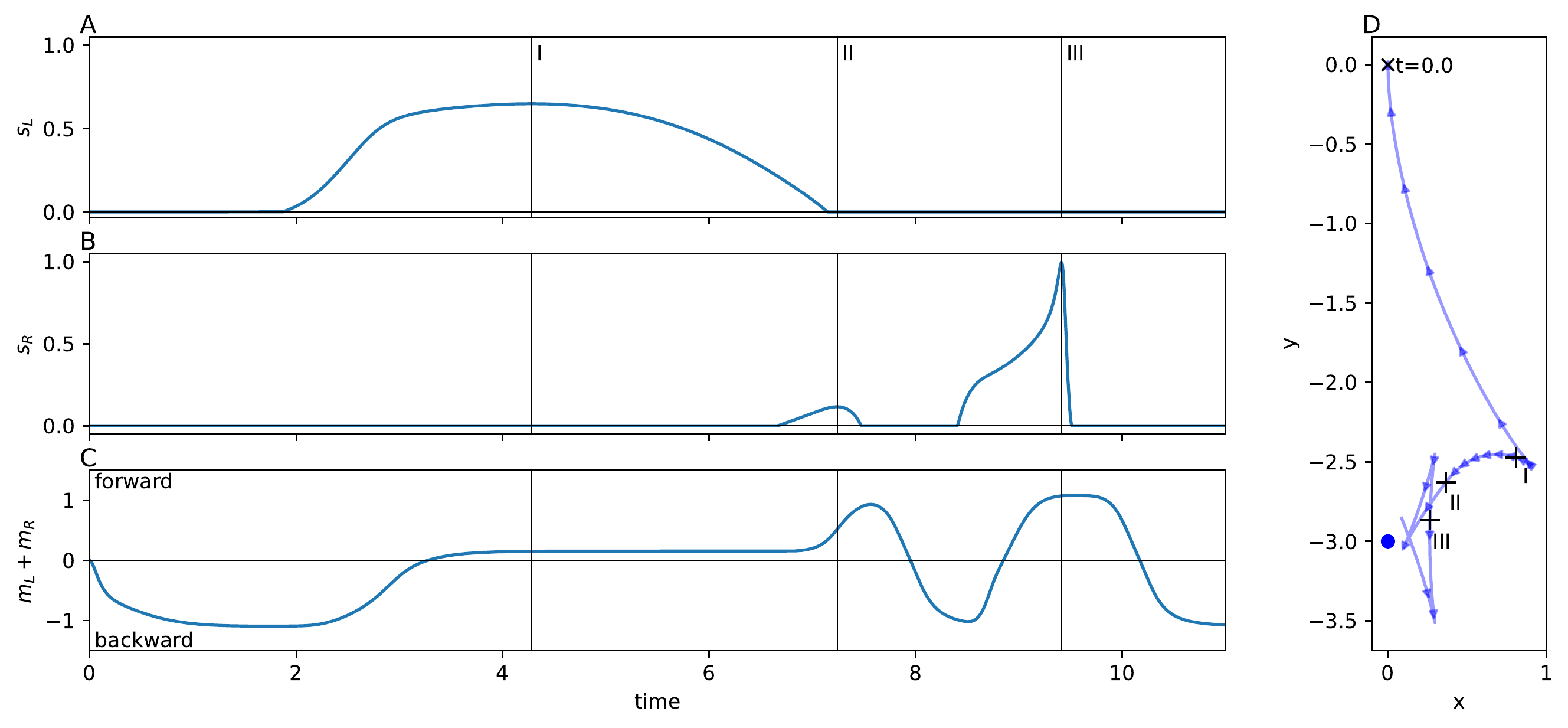}
\end{center}
\caption{
    A-C show the sensorimotor activity of the ancestral solution when the light is at coordinates (0, -3), position 6 in Figure \ref{fig_ancestor_5k_trajectories}.
    D plots the spatial trajectory of the robot.
    The vertical lines in plots A-C show the peaks in sensor activity.
    These correspond to the + markers in D.
    Initially the robot drives backwards.
    The left sensor stimulation between $t=2$ and $t=8$ is associated with the robot to driving forwards while turning strongly to the left.
    Once this turn has oriented the robot such that the right sensor is being stimulated and the left sensor is no longer being stimulated, the robot drives forward until the right sensor is no longer stimulated.
    From here, this is just the same Phase B behaviour presented in Figure \ref{fig_ancestor_detail_right}.
}
\label{fig_ancestor_detail_left}
\end{figure}

The approach phase (Phase A) often consists of simply driving forwards, and then continuing to drive forwards until the right sensor is not stimulated.
Thereafter the procedure for Phase B is followed, with the approach differing from  the orbit primarily in that that the amount of time spent on each step of Phase B while approaching the light varies more than it does when the robot is stably orbiting the light.
This is what we see in trajectory shown in Figure \ref{fig_ancestor_detail_right}, and in all conditions when the left sensor is not stimulated.
However in conditions when the left sensor is stimulated during the approach phase, the left sensor is involved in guiding the robot into a state where Phase B takes over.
This can be seen in Figure \ref{fig_ancestor_detail_left}.

This solution is an instance of a more general robust strategy for performing phototaxis in this model, which can be summarised even more simply as:

\begin{itemize}
    \item If you don't see the light, drive backwards (it must be behind you).
    \item If you do see the light, drive forwards until you can't see it any longer.
\end{itemize}

The reason this does not result in just driving backwards and forwards along the same arc is that the robot turns a different amount when driving forwards vs when driving backwards.
The turn amount is determined by $m_R - m_L$, while the direction of travel is determined by whether $m_R + m_L$ is negative or positive.
When adjusting motor activity to change directions, it's trivial to also change the amount of turn.
Of course this general strategy is not a complete description of the robot's behaviour, the effect of sensor stimulation can be time dependent and differ for the left and right sensors.
Particularly during Phase A, the approach to the light, the exact trajectories taken by the robot depend on continually regulating the 2 independent motors' speed and direction of activity to perform both gradual turns and sharp changes in direction via 3 point turns with sufficient precision to reliably enter Phase B and maintain it.
However, we see this general strategy well preserved in populations descendent from this ancestral population as well as evolved independently in non-descendent populations.

To summarise, the ancestral solution takes advantage of the particular nature of its sensors, driving backwards so that the sensors are stimulated sharply. It adjusts its motor activity in response to this sharp stimulation in such a way that the stimulation is extinguished. This environmentally mediated negative feedback loop plays a critical role in stably remaining in close proximity to the light source.

\subsection{Experiment 2: Avoidable Interference}

The simplest possible interference would be adding a constant value to all the sensor inputs. However this would not depend on the system's motor activity. Therefore the first $\psi(m)$ that we model is threshold based interference, where interference is maximised when motor activation is above a threshold value, and $\approx 0$ elsewhere. To achieve this effect with a smooth function, we use a relatively steep sigmoidal function, with the equation:

\begin{equation}
    \psi(m) = \frac{1}{1+\exp({-k(|m|-p)})}
    \label{eq_sigmoid_interference}
\end{equation}

Where $\exp(x) = e^x$ and $|m|$ is the absolute value of $m$, and where $k=50$ is the term controlling the steepness of the sigmoid's transition from 0 to 1,
while $p=0.5$ determines the midpoint of the transition.
So when
$m < -0.5$
or $m > 0.5$:
$\psi(m) \approx 1$
and when
$-0.5 < m < 0.5$:
$\psi(m) \approx 0$.
This function is unique among the three in that its interference is completely avoidable by constraining the system's motor activity.
We will refer to the interference generated by this function as \textit{avoidable} or \textit{sigmoidal} interference.

\label{sec_results_sigmoid}

\begin{figure}[t]
\begin{center}
\includegraphics[width=0.95\textwidth]{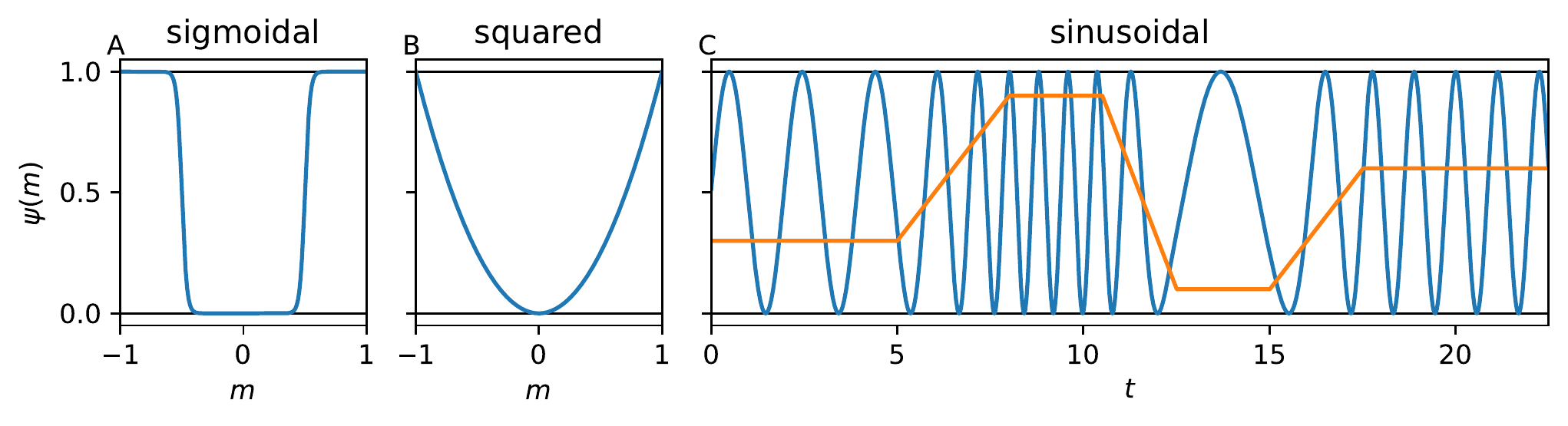}
\end{center}
\caption{
    A and B plot pure functions of $m$ corresponding to Equations \ref{eq_sigmoid_interference} and \ref{eq_squared_interference}.
    C plots a function of time that depends on the cumulative history of $m$, Eq. \ref{eq_sinusoidal_interference}.
    The blue line is the interference, while the orange line is the motor activity.
}
\label{fig_interference_functions}
\end{figure}

With motor activity capped at 50\%, motor-driven interference can be avoided, and phototaxis can still be performed, just more slowly.
Moving more slowly comes at a cost to fitness though, since the fitness function (Equation \ref{eq_fitness_func}) rewards reaching the light quickly.
Therefore, a predict-and-subtract solution to the interference which preserves the speed of the high-performance ancestral solution should outperform a solution which simply avoids the interference.
However we instead found that the fittest solution from the 5 populations evolved to perform phototaxis with sigmoidal interference function modifies the motor activity of the ancestral solution significantly.

\begin{figure}[t]
\begin{center}
\includegraphics[width=\textwidth]{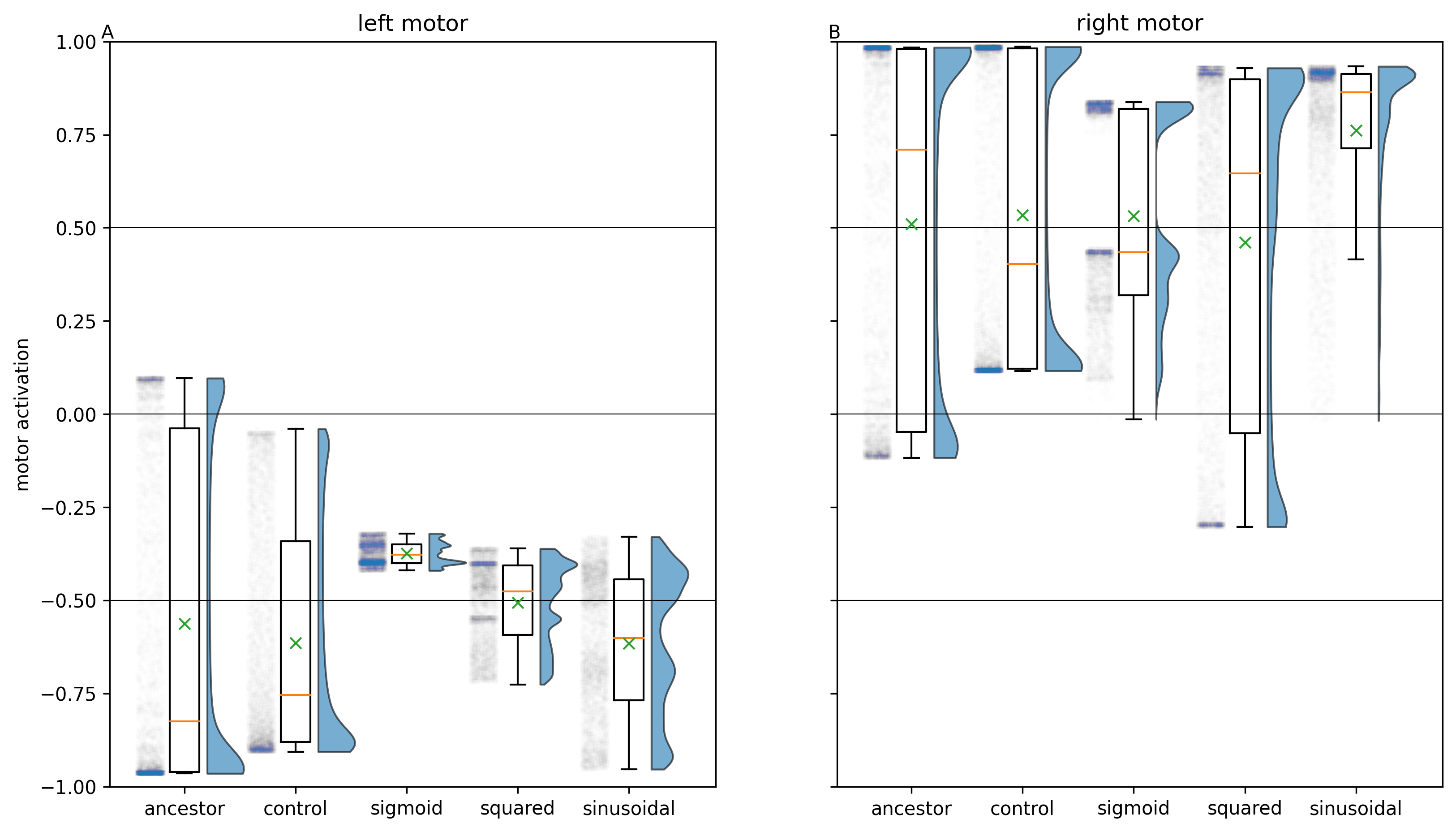}
\end{center}
\caption{
    Motor activity for the 12 light positions shown in Figure \ref{fig_ancestor_5k_trajectories} for time 20 to time 50 (integration steps 2000-5000), for the evolved solution to each experiment.
    The boxes extend from the first to the third quartile of the motor activity,
    and contain a yellow line showing the median,
    and a green $\times$ showing the mean.
    The whiskers extend to 1.5 times the inter-quartile range.
    The half-violin plot to the right of each box plot estimates the distribution of the motor activity,
    while to the left is a scatter plot of each simulated moment of motor activity with randomized horizontal placement.
    The column labelled control plots exactly the same information for the fittest solution evolved with $\lambda=0.5$ and the null interference function $\psi(m) = 0$,
    showing the scope of change seen simply due to the presence of $\lambda$ and to genetic drift.
}
\label{fig_motor_raincloud}
\end{figure}

\begin{figure}[t]
\begin{center}
\includegraphics[width=\textwidth]{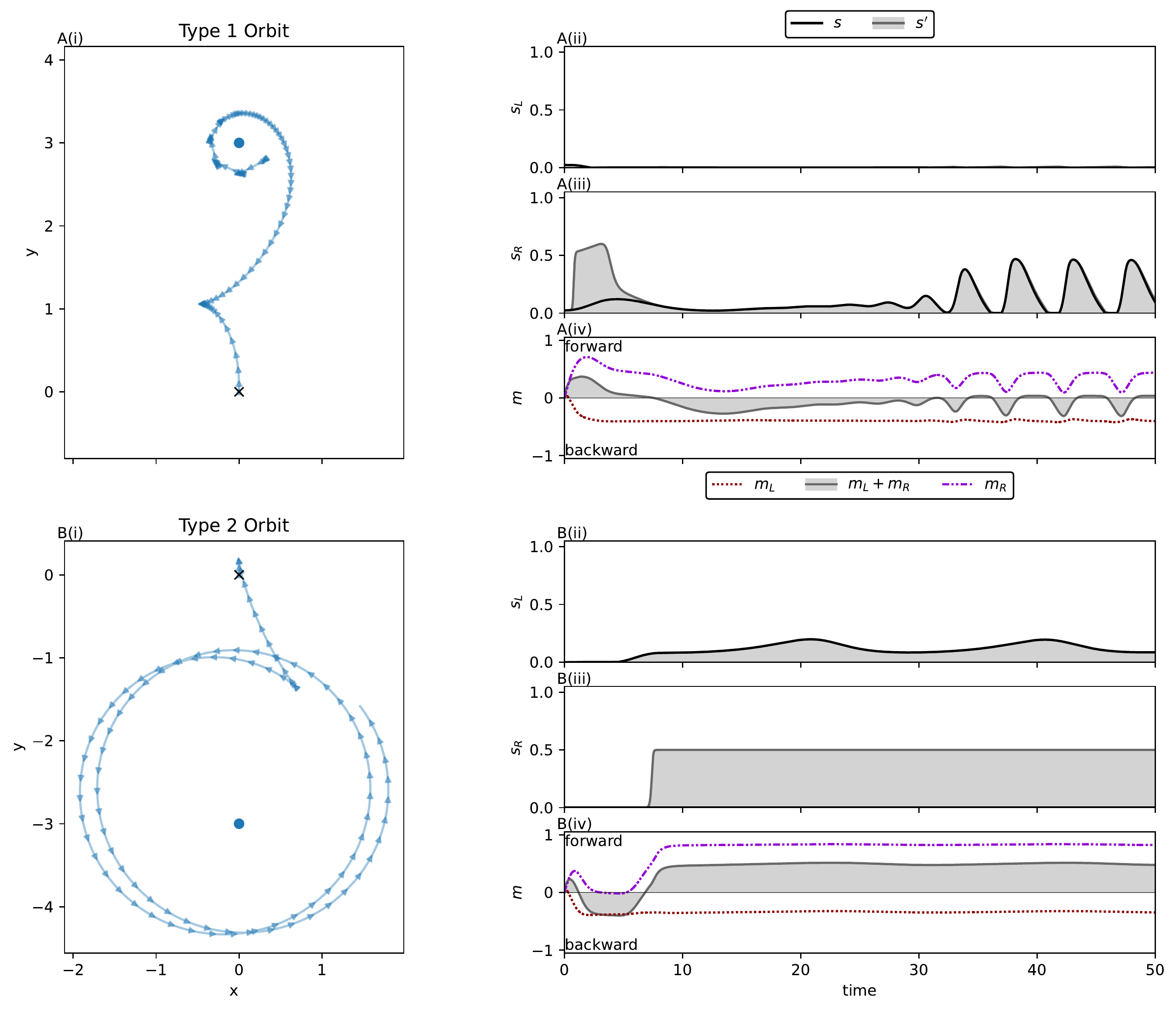}
\end{center}
\caption{
    Two distinct orbit types produce the bimodal right motor activity distribution seen for the solution evolved with sigmoidal interference in Figure \ref{fig_motor_raincloud}B.
    Note the minimisation of interference during the Type 1 orbit, in contrast with high level of right sensor interference during the Type 2 orbit.
}
\label{fig_timeseries_sigmoid}
\end{figure}

Figure \ref{fig_motor_raincloud} illustrates how the characteristic motor activity of the solution evolved with sigmoidal noise differs from that of the ancestral solution.
Keeping in mind that the ancestral solution often involved minimal environmental stimulation of the left sensor,
we observe that the left motor in this evolved solution never produces interference.
This comes at the cost of greatly decreased absolute motor activity relative to the ancestral solution.
The ancestral solution's left motor activity ranges widely, from -0.96 to 0.10 with a median of -0.82, close to the maximum possible absolute value of 1.
See Figure \ref{fig_ancestor_detail_right}A(ii) for ancestral motor activity as a time series.
In contrast, the left motor activity of this solution ranges only between -0.42 and -0.32 with a median value of -0.38.
Time series of this motor activity can be seen in Figures \ref{fig_timeseries_sigmoid}A(iv) and \ref{fig_timeseries_sigmoid}B(iv).
This drastic decrease in motor activity lowers the speeds attainable by the robot, but prevents motor-driven interference with the left sensor.
While the activity of the left motor is kept below the threshold for producing interference at all times, keeping the left sensor free of interference, the right motor does produce interference.
The distribution of right motor activity is bimodal, with peaks just below the interference threshold of 0.5, and close to its maximum value of 0.84.
This bimodal distribution is the result of this solution producing two distinctly different orbit types.

\begin{figure}[t]
\begin{center}
\includegraphics[width=0.8\textwidth]{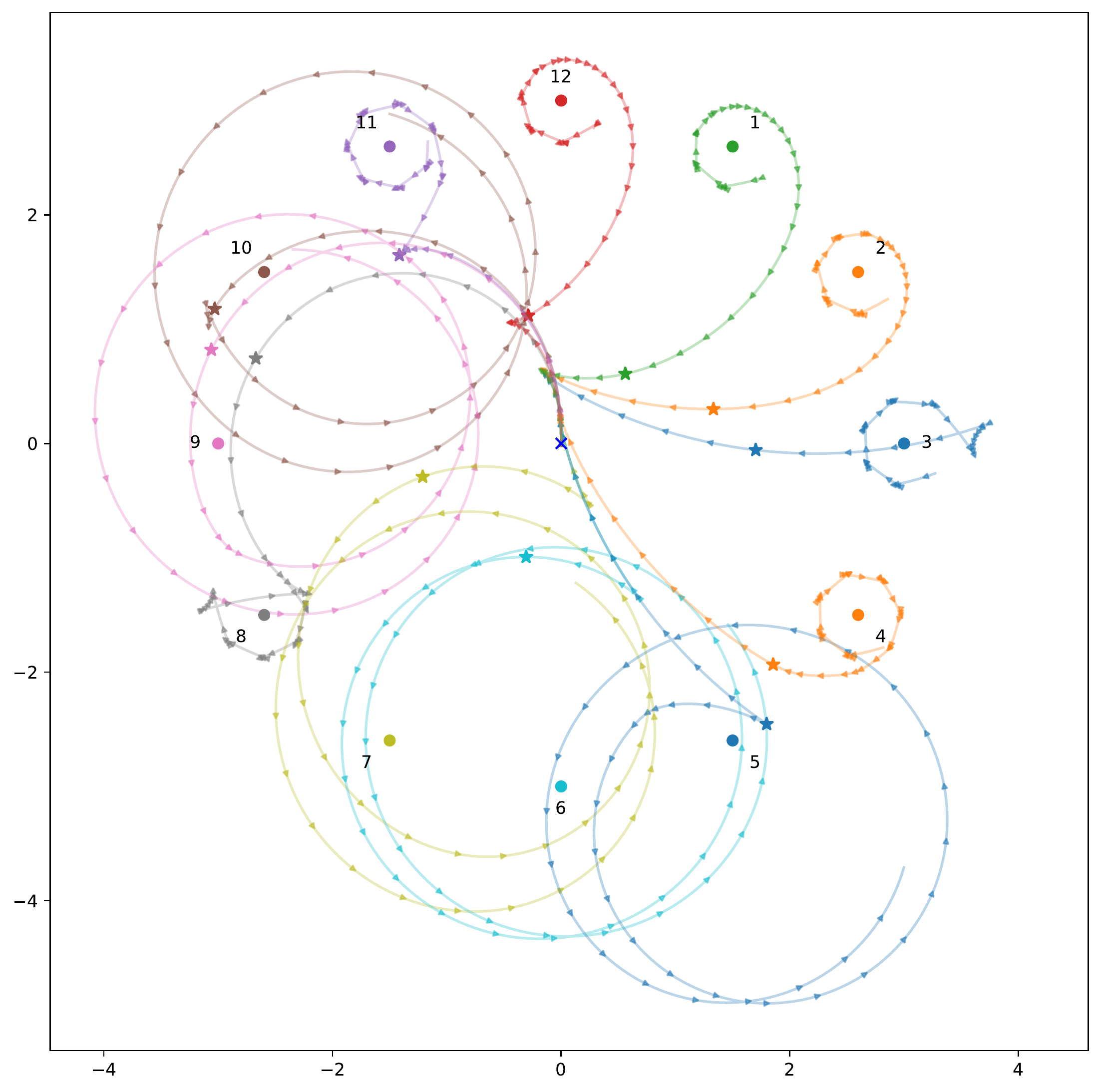}
\end{center}
\caption{
    Two distinct types of orbits are visible in the 
    spatial trajectories for the best individual from populations evolved with sigmoidal interference (Eq. \ref{eq_sigmoid_interference}).
    Type 1 orbits, reminiscent of the ancestral solution, are seen for Lights 11-4.
    Type 2 orbits, which feature a forward moving, counter-clockwise orbit of the light are seen for Lights 5-10, with the exception of Light 8, where the approach puts the robot in position for a Type 1 orbit.
}
\label{fig_sigmoid_trajectories_5k}
\end{figure}

The orbiting behaviours of this system are of interest because they demonstrate ways in which a long term, stable relationship with an environmental source of sensor stimulation can be maintained in a model with motor-driven sensor interference.
Like for the ancestral population, a trial duration of 10 time units was used for this population.
Due to the decreased overall motor activation relative to the ancestor, and the consequently decreased speed, the robot does not get as close to the light in that time as the ancestor did.
This means that what has been selected for by the genetic algorithm here is modification of the approach phase to maintain accuracy in the presence of this novel interference.
However, due to a sufficiently accurate approach and the evolved regulation of the motor-driven interference, stable orbits are still achieved across all light positions in the very long term.
Unlike the ancestor, we see two distinctly different orbit behaviours.
Across all interference functions we refer to those orbits reminiscent of the ancestral solution, involving forward and backward motion around the light, as Type 1 orbits,
and to orbits which loosely circle the light while driving forwards as Type 2 orbits.
These are easily distinguished visually (see Figure \ref{fig_sigmoid_trajectories_5k}).
As with the ancestor, approaches can broadly be divided into those guided by the left sensor, and those that are not.
In the majority of cases for this solution, the approach phase preceding Type 1 orbits is guided exclusively by the right sensor,
while Type 2 orbits tend to follow a left sensor guided approach phase.

Type 1 orbits come much closer to the light.
They display similar sensorimotor behaviour to the ancestor's orbit behaviour (Phase B), maintaining a stable relationship to the light by repeatedly driving backwards and forwards, albeit with greatly reduced motor activity compared to the ancestor.
Figure \ref{fig_timeseries_sigmoid}A shows a typical example of sensorimotor activity for Type 1 orbits.
Right motor-sensor interference is almost entirely avoided.
A very low amount (not visible in the figure) coincides with the robot driving forwards slowly.
This interference is necessary because the left motor's activity is negative, and is maintained very closely to the threshold for interference,
so the right motor's positive activity cannot be raised sufficiently highly to drive forwards without producing at least a small amount of interference.
We summarise this orbit strategy as performing the known good ancestral strategy while constraining motor activity to avoid sensor interference.

Type 2 orbits loosely circle the light, and are very different from the ancestral orbit behaviour.
Figure \ref{fig_timeseries_sigmoid}B shows an example of typical sensorimotor activity for this type of orbit.
These orbits do not involve environmental stimulation of the right sensor, instead the left sensor is stimulated throughout the orbit phase.
Unlike Type 1 orbits, where the relationship to the light is maintained by repeatedly driving forwards and backwards, the robot exclusively drives forwards.
It does so very quickly, producing high right motor-sensor interference.
We characterise this orbit strategy as keeping `one eye on the prize', where the left sensor, facing the light, is kept free of interference.
Meanwhile the right sensor, facing away, is continually stimulated by the right motor's activity.
This orbit strategy is uniquely enabled by the ipsilateral nature of the motor-driven sensory interference.

In the presence of this threshold based interference, the best solution found by our GA when modifying the ancestral population to accommodate this interference constrains the ancestral solution's motor activity to avoid interference while performing the same function of phototaxis, using (in some situations) the same basic strategy.
This approach contrasts with the predict-and-subtract approach of modifying the controller to subtract the anticipated interference from the sensor neurons' outputs, allowing the behaviour of the ancestral solution to be performed without modification.
This suggests that in our model such solutions are far closer in evolutionary space to the ancestral solution than a predict-and-subtract solution would be.
The relevance of this to the evolutionary history of biological control systems is unclear, however it may suggest that adjusting neural activity to accommodate a novel form of motor-driven sensory interference would involve regulation of the behaviour producing that interference in addition to or instead of the subtraction of predicted interference.
This demonstrates that
behavioural modification does indeed work as a solution to motor-driven sensory interference, and
that the precise way in which behaviour is modified can depend heavily on the particularities of the sensorimotor contingency in question.
Specifically we have seen how two ways of compensating for motor-driven sensory interference emerged in our model.
Firstly, motor activity may be constrained to ranges that minimise or avoid interference with the sensors.
Secondly, interference can be avoided for only one sensor, which is kept trained on relevant environmental stimuli. This permits unconstrained use of motor activity which interferes with the other sensor.

\subsection{Experiment 3: Unavoidable Interference}
\label{sec_results_squared}

Sigmoidal interference certainly does not exhaust the possibilities for modelling interference, nor does it capture the fact that many self-caused stimuli cannot be avoided when taking action.
Therefore we also model non-avoidable interference, where the interference increases with the absolute magnitude of the motor activation. To minimise discontinuities in the system, we use the square of the motor activity:

\begin{equation}
    \psi(m) = m^2
    \label{eq_squared_interference}
\end{equation}

We will refer to the interference generated by Equation \ref{eq_squared_interference} as \textit{unavoidable} or \textit{squared} interference.
Like the threshold based interference modelled previously, the magnitude of the interference correlates with the magnitude of the motor activity.
However, now all changes in motor activity produce a corresponding change in the sensory interference.

\begin{figure}[t]
\begin{center}
\includegraphics[width=0.8\textwidth]{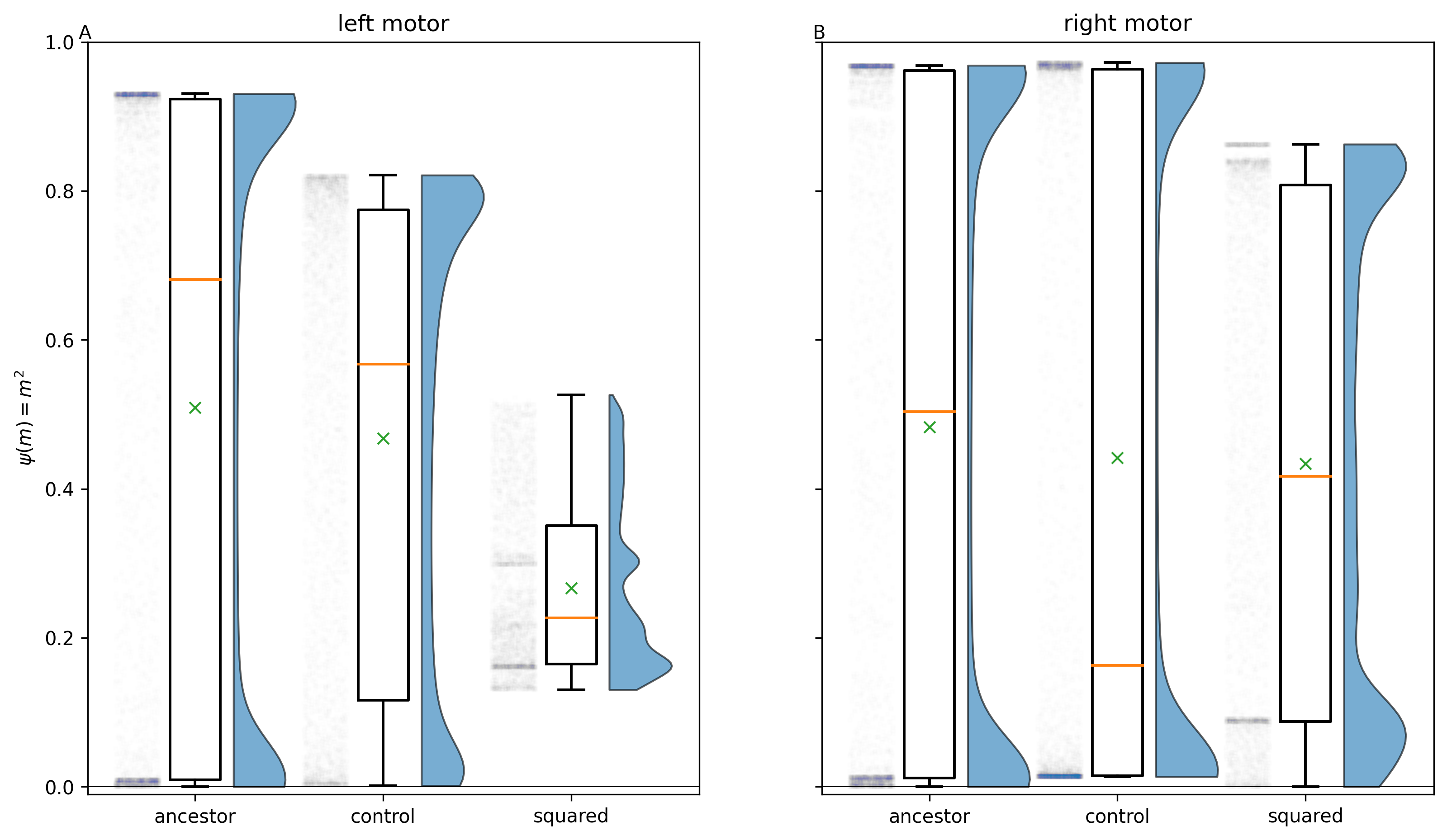}
\end{center}
\caption{
    Motor-driven interference is reduced in Experiment 2 relative to the ancestral population.
    The figure shows $\psi(m) = m ^ 2$ for the 12 light coordinates shown in Fig. \ref{fig_ancestor_5k_trajectories}, for $20 < t < 50$.
    Note primarily the lowered mean, median and maximum interference with the left motor.
    Despite the right motor's activity being spread across a wider range than either ancestor or control (see Figure \ref{fig_motor_raincloud}), this spread is to low motor activity values, decreasing maximum right motor-sensor interference.
    However the right motor activity has definitely not been suppressed the way the left has.
}
\label{fig_sq_interference_comparison}
\end{figure}

\begin{figure}[t]
\begin{center}
\includegraphics[width=0.7\textwidth]{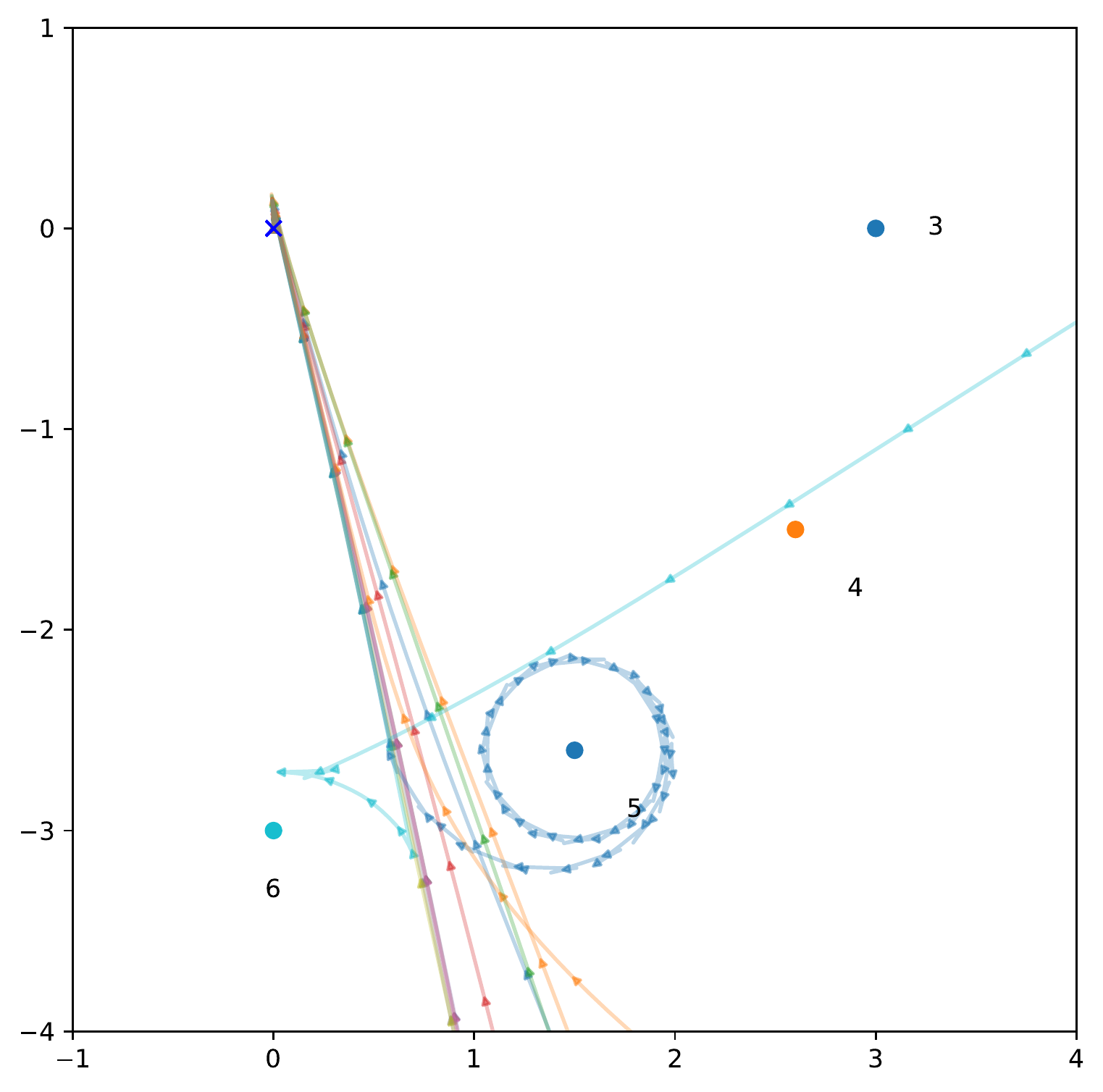}
\end{center}
\caption{
    When motor-driven interference is removed, the behaviour evolved with squared interference fails.
    Spatial trajectories for 12 light coordinates (Fig. \ref{fig_ancestor_5k_trajectories}) are plotted with all motor-sensor interference removed.
    The approach phase now only succeeds in two out of 12 cases, where the blind approach brings the robot close to the light.
    The orbit phase only succeeds in one of these two cases.
}
\label{fig_sq_minus_l_interference}
\end{figure}

\begin{figure}[t]
\begin{center}
\includegraphics[width=\textwidth]{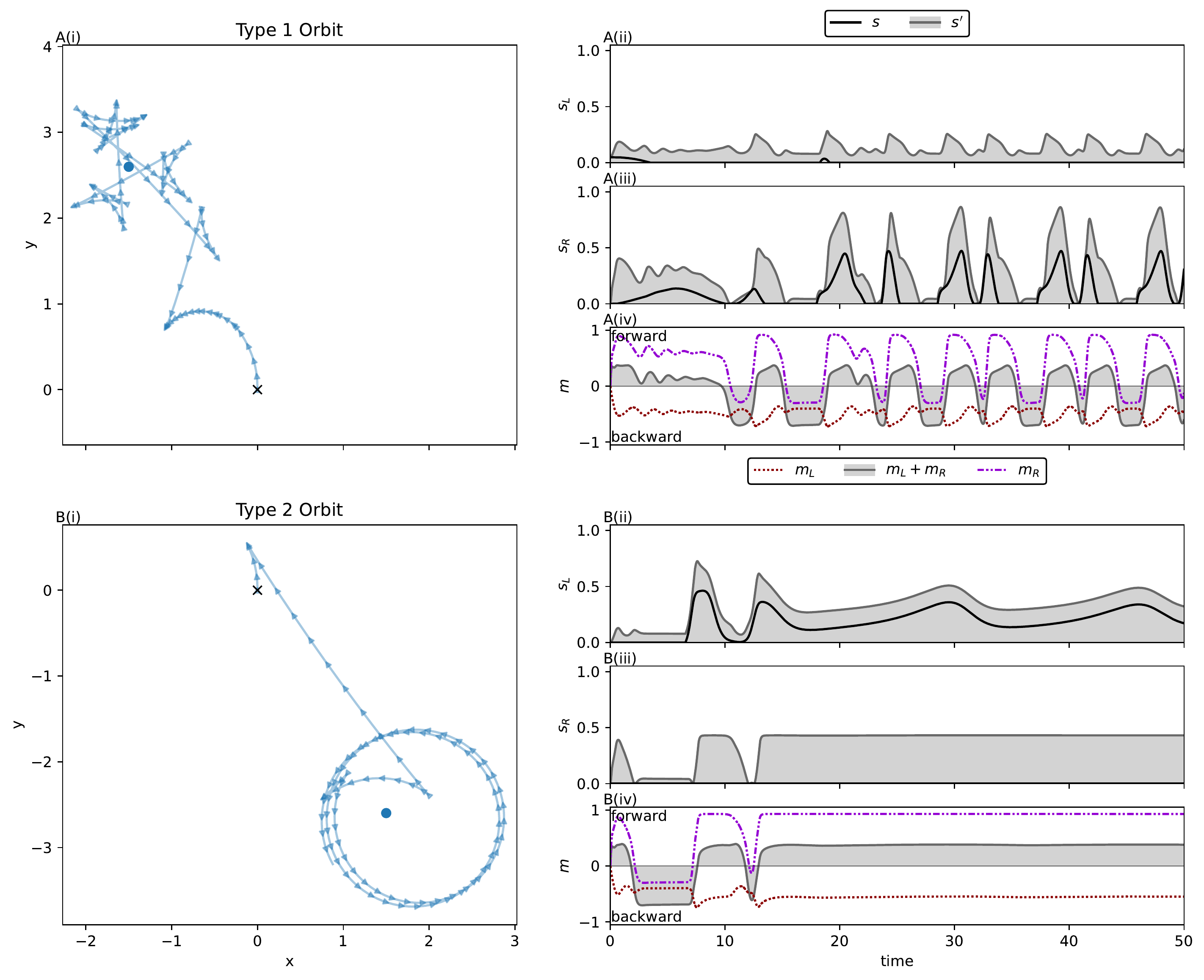}
\end{center}
\caption{
    Spatial trajectories and sensorimotor activity showing a Type 1 and Type 2 orbit for the solution evolved with squared interference. 
    A shows a Type 1 orbit reminiscent of the ancestral solution,
    where motor activity is coordinated with sharp spikes of environmental stimulation of the right sensor.
    A(iii) shows how the right motor interference amplifies the environmental stimuli.
    The spiking activity is characteristic of negative feedback in this solution, where action resulting from sensor stimulation leads to the stimulus diminishing.
    B shows a Type 2 orbit, where the robot orbits while driving forwards.
    B(iv) shows how the motor activity plateaus during the orbit, with high right motor interference seen in B(iii).
    This is associated with positive feedback in this solution,
    where sensor stimulation leads to activity prolonging that stimulation.
}
\label{fig_timeseries_squared}
\end{figure}

Examining the fittest solution produced by the GA's modification of the ancestral solution, we again find the ancestral solution well preserved.
A trial duration of 20 time units was used during evolution to compensate for any decreased speed compared to the ancestor.
The general strategy of approaching the light while driving backwards is maintained, however motor activity has changed to accommodate the addition of the squared interference function.
The left motor's activity is now constrained to a much smaller range (see Figure \ref{fig_motor_raincloud}A), which lowers interference dramatically compared to the interference that would be produced by the ancestral solution's motor activity (see Figure \ref{fig_sq_interference_comparison}A).
The right motor generates significant interference, but we find that rather than destructively interfering with the sensor in such a way that the environmental stimuli is masked,
this motor-driven sensor stimulation is actually constructive in that it synchronises with and amplifies the environmental stimuli's effect on the sensor.
Figure \ref{fig_motor_raincloud}B makes it clear that the right motor's activity has not been lowered or even constrained to a tighter range the way the left motor's has - though we still see a slight reduction in interference compared to what the ancestral solution would produce (see Figure \ref{fig_sq_interference_comparison}B).
How the system performs so accurately in the presence of this interference becomes clear when we consider the relationship between the right motor activity and the right sensor.
As with the ancestor, the robot approaches the light while driving backwards, in such a way that the light enters the right sensor's field from it's blind spot at very close proximity to the sensor.
Figure \ref{fig_timeseries_squared}A shows an example of this approach.
When the light enters the right sensor's field, its activation immediately spikes.
In response, the right motor's activity also spikes, causing the robot to drive forwards, and also causing a spike of interference in the same sensor.
This is a version of the ancestral Phase B orbit behaviour, executed with reduced baseline motor activity, and high right motor activity coordinated with right sensor stimulation.
By keeping motor activity at a low baseline and interacting with the environment in such a way that environmental stimuli are sharp and intense, this solution facilitates distinguishing environmental stimuli from low levels of self-caused background noise.
By then coordinating activity with environmental stimuli,
motor-driven interference can be raised to high levels without interfering with the system's function,
`hiding' in the shadow of the environmental stimuli.
Not only does this activity not interfere with perception of the environment,
the stimulation caused by right motor's activity actually reinforces and amplifies the environmental stimuli's effect on the sensor above the maximum level it would be able to achieve on its own.

\begin{figure}[t]
\begin{center}
\includegraphics[width=0.8\textwidth]{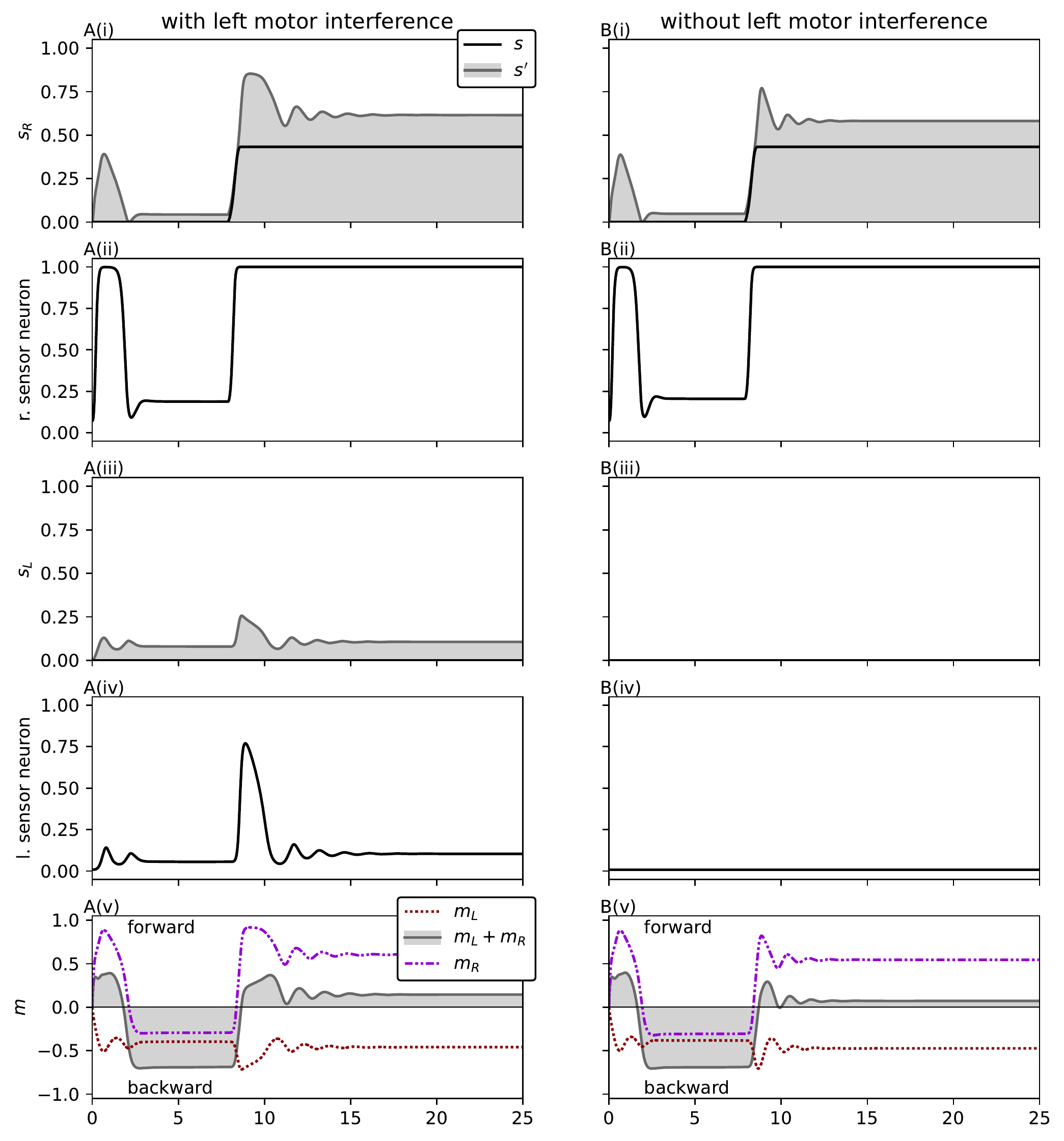}
\end{center}
\caption{
    The magnitude and duration of the initial motor response to sensor stimuli are strengthened by the presence of left motor interference.
    Sensorimotor activity and sensory neuron output time series are shown for the solution evolved with squared interference (Eq. \ref{eq_squared_interference}),
    when the right sensor is presented with an artificial environmental stimuli, which spikes and plateaus around $t=8$.
    A shows the response under the condition of evolutionary adaptation for the robot, with motor interference present.
    B shows the response when the left motor-sensor interference is removed.
    The duration and intensity of the motor response to the stimuli is diminished without the interference, indicating that the interference plays a functional role in the evolved behaviour.
    Additionally, it can be seen that the response to sudden right sensor stimulation is accompanied by internal negative feedback - even when the stimulation persists, motor activity quickly falls from the initial peak.
}
\label{fig_squared_right_spike_and_plateau}
\end{figure}

Since right sensor stimulation leads to right motor activity, which in turn leads to more right sensor stimulation, we should address the possibility of a self-sustaining positive feedback loop.
This possibility is limited by two forms of negative feedback.
The system's relationship to the light source is structured in such a way that elevated right motor activity in response to the environmental stimuli moves the right sensor away from the light, eliminating that stimuli.
This is environmentally mediated negative feedback.
It is complimented by internal negative feedback.
Figure \ref{fig_squared_right_spike_and_plateau}A shows how a spike in right sensor stimulation causes an initial strong response in motor activity.
However despite continued stimulation at an elevated level, sufficient to saturate the output of the sensor neuron, motor activity quickly falls from the initial peak. 
Thus both internal and environmentally mediated negative feedback play a role in preventing this orbit behaviour from being disrupted by positive feedback.

As we also saw with sigmoidal interference, this solution realises a second orbit pattern of Type 2.
Positive feedback instead plays a role in this orbit,
which comes into effect when the robot is close to the light, but the light is on its left (see Figure \ref{fig_timeseries_squared}B).
The system's response to left sensor stimulation does not feature the internal negative feedback that right sensor stimulation does, and it produces a response in both right and left motor activity.
This in turn produces interference in both sensors.
The ultimate effect is that the robot drives forwards in a counter-clockwise orbit around the light.
This keeps the left sensor continually stimulated by the light, while the right sensor is continually stimulated by the right motor's activity.
In this case we have an environmentally mediated, positive feedback loop, where left sensor stimulation causes the robot to turn towards that stimulus, and the motor-sensor interference produces the same effect.

The way this system has been parametrised relies on the presence of motor-driven stimulation to perform phototaxis.
Recall that the ancestor evolved to have zero left sensor activation in many situations, with a left sensor guided approach phase (Phase A) for a number of initial light positions.
This trait remains in a way, where the left sensor is often completely free of environmental stimulation, and the left motor activity is constrained to produce lower levels of interference.
However this interference plays an important role.
Figure \ref{fig_sq_minus_l_interference} illustrates how removing the motor-driven sensor stimulation from the left sensor causes the approach phase to fail in the majority of cases, succeeding only when its trajectory inadvertently brings it close to the light.
This is not unexpected, given that the system was optimised for the presence of motor-driven interference.
However it means that accurate control of the system's motor activity has been optimised in such a way that it now depends on perceiving the direct sensory effects of its own activity.
Like the right motor, the left motor responds to sensor stimuli,
though in a smaller range and with elevated negative rather than positive activation.
This plays an interesting role in the system's response to right sensor stimulation (as in the Type 1 orbit shown in Figure \ref{fig_timeseries_squared}A).
Note how the coordinated peaks of right environmental and motor-driven sensor stimulation coincide with elevated left motor activity and corresponding motor-driven left sensor stimulation.
Figure \ref{fig_squared_right_spike_and_plateau} shows how the presence of left motor-sensor interference amplifies and extends the initial motor activity response to right sensor stimulation.
This demonstrates not only a specific way in which the system has been optimised for the presence of interference, but also how self-caused stimuli can play a directly functional role in behaviour.

To summarise, we see the ancestral strategy is well preserved in this evolved solution.
This solution can be characterised as minimising interference to an extent,
as we also saw in the case of sigmoidal interference.
We also see a condition where motor-driven sensor interference does not need to be minimised, namely when it can be made to coincide temporally with environmental stimulation of the same sensor.
Here the onset of the environmental stimuli prompts the interfering motor activity, and a combination of internal and environmentally mediated negative feedback extinguishes both interfering activity and stimuli.
In this case the motor-driven stimulation does not interfere with perception of the environmental stimuli, instead reinforcing and amplifying it.
This obviates the need to distinguish or subtract the self-caused stimuli from the environmental.
Separately, we also see that a stable, periodic orbit phase can be facilitated by positive feedback.
Finally, we found that while left motor-sensor interference is confined to a narrow range,
the system has been optimised to rely on its presence and even incorporate it functionally.

\subsection{Experiment 4: Time Dependent Interference}
\label{sec_results_sin}

With both of the preceding interference functions, if the motor activity is held constant, then the interference will also take on a constant value.
Since the interference is additive and non-saturating,
subtracting a constant term can remove the interference and leave only the environmental signal.
In general a CTRNN with a sufficiently high bias $\beta$ for the input neurons can do this.
Solutions to the previous two interference functions have shown both the utility of avoiding or minimising motor-sensor interference,
as well as the role that holding motor activity and its corresponding interference constant can have in constructing long-term stable relationships with sources of environmental stimuli.
With the following function it is not possible for the interference to plateau at a constant value.
It describes a sine wave with a maximum of 1 and a minimum of 0, whose frequency is determined by the motor activation:

\begin{alignat}{2}
    \psi &= \frac{\sin(c) + 1}{2}
    \label{eq_sinusoidal_interference}
    \\
    \dot{c} &= (b + |m|)r 
\end{alignat}

Here $c$ gives the phase of the sinusoidal, capturing the previous values of $m$.
$b=0.1$ determines the base frequency of the sinusoidal in the absence of any motor activity,
while $r=8$ is the frequency range term determining the maximum frequency the sinusoidal can reach.
The effect of adding 1 and dividing by 2 is simply to shift the wave from the range $[-1, 1]$ to the range $[0, 1]$.
This equation essentially advances through a standard sine wave at a rate determined by the motor activity.
As with the previous interference functions, the interference for a given sensor is calculated from the ipsilateral motor, such that when computing the interference for the left sensor we have $m = m_L$, and for the right sensor $m = m_R$.

\begin{figure}[t]
\begin{center}
\includegraphics[width=\textwidth]{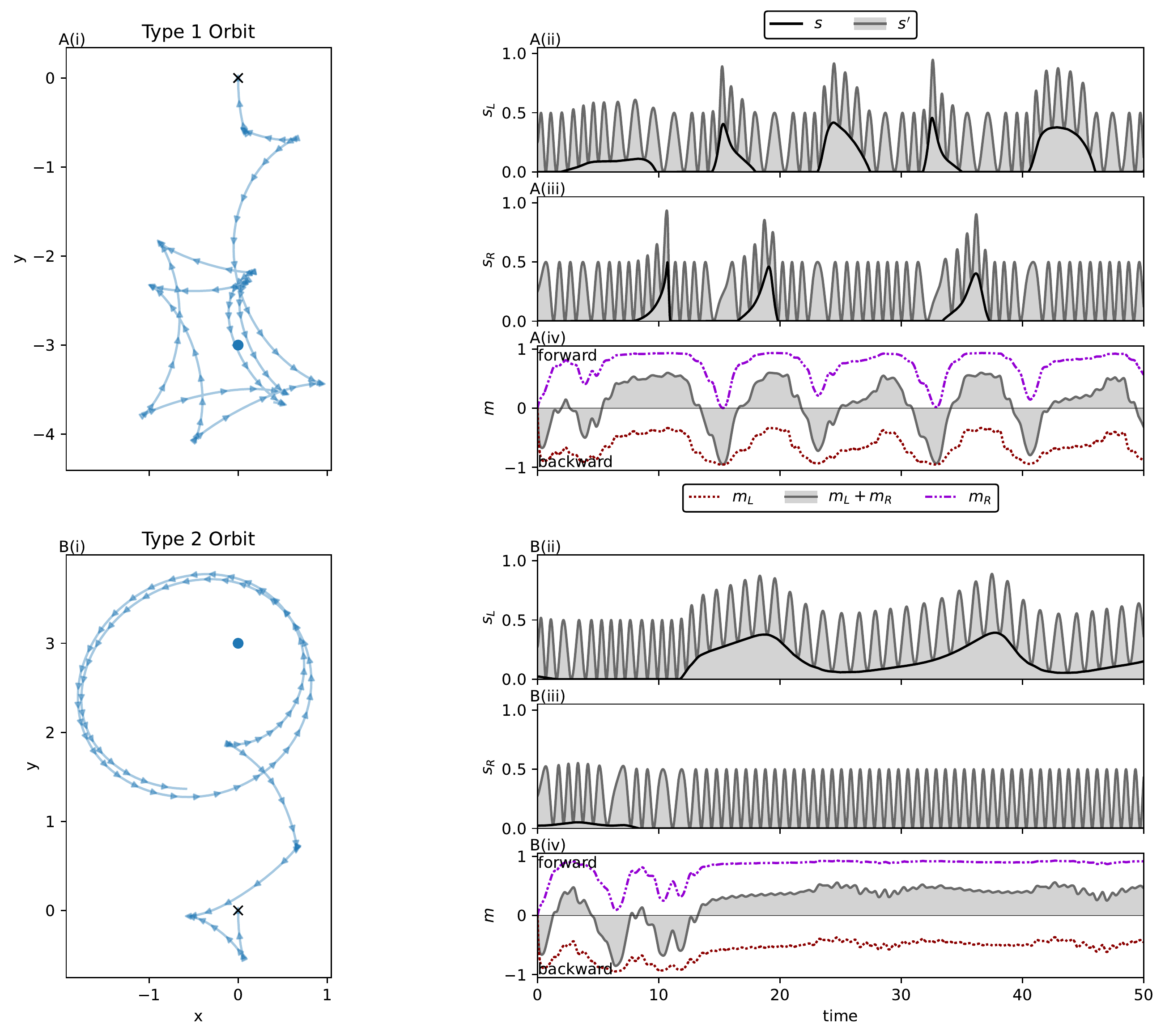}
\end{center}
\caption{
    Spatial trajectories and sensorimotor activity
    for the solution evolved with sinusoidal interference.
    The sensor plots show how the relatively slowly changing environmental sensor stimulation raises the minima of the high frequency interference, allowing the environmental stimuli to be responded to despite the interference.
    The difference in timescale that makes this possible is clearly visible here.
    Responsiveness to the environment is most clearly visible in A(iv), where more positive motor activity is associated with environmental stimulation of the left or right sensor.
    The continual oscillations in motor activity (most clearly visible in the grey net motor activity line) are driven by the high frequency interference.
    These oscillations produce the elliptical Type 2 orbit seen in B(i).
}
\label{fig_timeseries_sinusoidal}
\end{figure}

Unlike the previous interference functions, this is not purely a function of the motor activity, such that if you know $m$ at time $t$, you know $\psi$ at time $t$.
Instead it is a function of time, depending on the prior history of the system, specifically on all the previous motor activity up to the current time.
More importantly for our purposes, if the input is held constant, the output continues to vary over time.
We will refer to the interference generated by Equation \ref{eq_sinusoidal_interference} as \textit{time dependent} or \textit{sinusoidal} interference.
A trial duration of 20 time units was used during evolution for this interference function.

Using this time dependent interference function we find that while avoiding interference, minimising it, or holding it constant are all important ways of coping with self-caused stimuli, they are not the only ways.
Timescale differences between the frequency of the motor-driven interference and the frequency of environmental stimulation of the sensor can be exploited to distinguish the two,
and behaviour can shape both interference and environmental stimuli to amplify these differences.

\begin{figure}[t]
\begin{center}
\includegraphics[width=\textwidth]{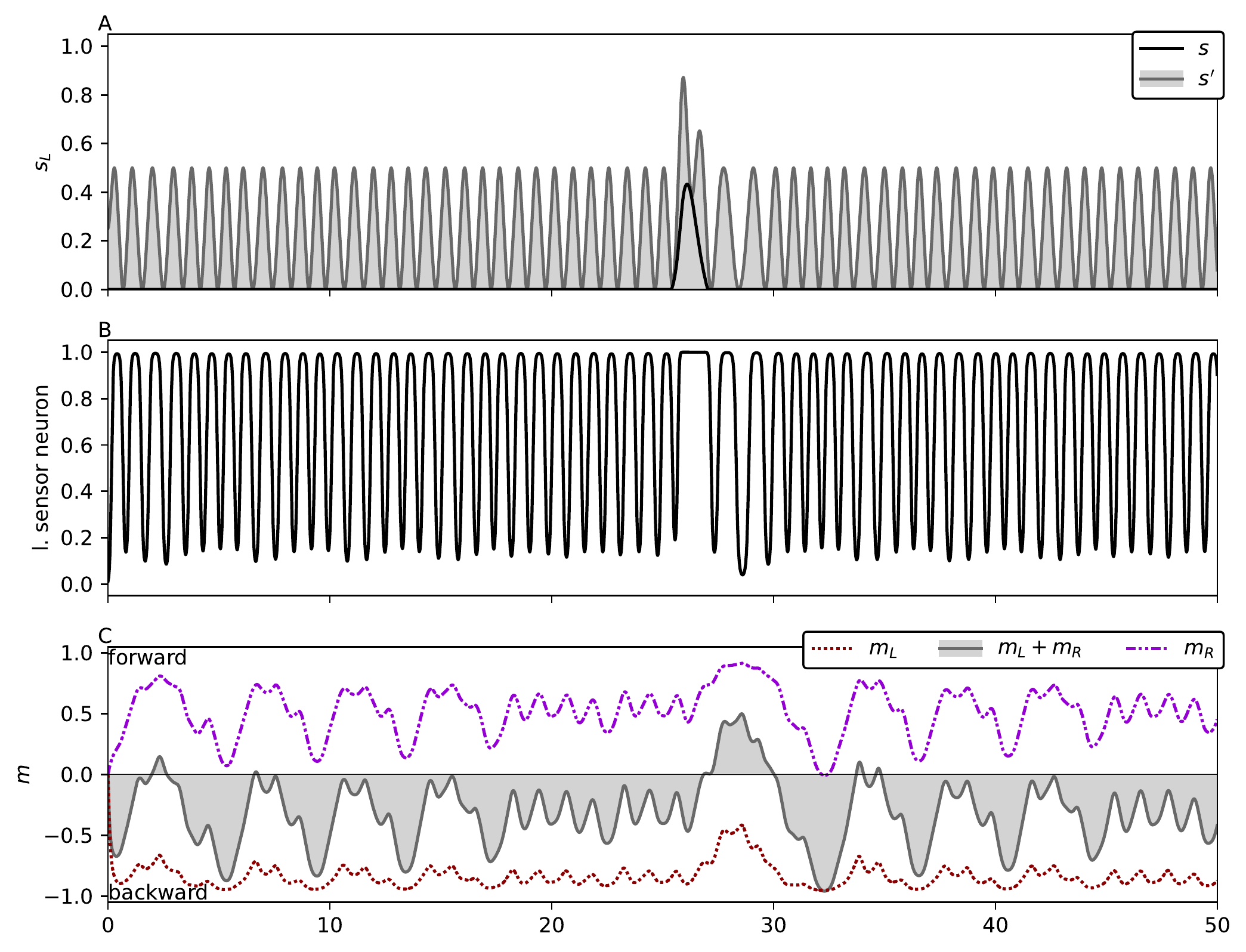}
\end{center}
\caption{
    Sensorimotor activity and sensory neuron output time series are shown for the solution evolved with sinusoidal interference (Eq. \ref{eq_sinusoidal_interference}),
    when the left sensor is presented with a spike in environmental stimulation at around $t=28$.
    The spike of environmental sensor stimulation causes the robot to drive forward instead of backwards for a time, demonstrating that the system can respond to environmental stimuli.
    The neural response to environmental stimuli is clearly visible - prolonged saturation of the left sensor neuron's output function (see Eq. \ref{eq_ctrnn}).
}
\label{fig_sinusoidal_late_spike}
\end{figure}

In this system the environmental signal is able to be detected despite the presence of interference,
due differences in timescale between the motor driven interference and typical encounters with the environmental stimuli.
First let's demonstrate that the system actually can respond to environmental stimuli. 
Figure \ref{fig_sinusoidal_late_spike} illustrates how
a spike in environmental stimulation of the left sensor has an excitatory effect on both motors, causing the system to switch from driving backwards to driving forwards.
Observing the behaviour of the output functions of this system's two sensor neurons, we found elevated neural biases $\beta$ compared to the ancestral solution: 4.67 and 3.73 for the left and right motor respectively compared to -0.75 and 0.99 in the ancestral solution.
These sensor neuron biases are calibrated such that
(A) with no environmental stimulation, the neuron's output function is maximised only with the peaks of the sinusoidal interference,
and (B) when combined with sufficient environmental stimulation, the troughs of the sinusoidal interference are high enough that the output function is maximised continually.
This can be seen in the neural response to environmental stimulation shown in Figure \ref{fig_sinusoidal_late_spike}B.
This makes the environmental signal detectable despite the continuously varying interference.
This solution is made possible by the large difference in timescale between the frequency of the sinusoidal interference and the frequency with which the sensor receives the environmental stimulation.
The frequency of the interference can be an order of magnitude higher, as can be seen in Figure \ref{fig_timeseries_sinusoidal}.
This difference in timescale means that the minimum value of the sinusoidal interference is bound to coincide multiple times with each period where there is no environmental sensor stimulation.
This means that a drop in neuron firing always coincides with the absence of environmental sensor stimulation, so over time the system can reliably respond to environmental stimuli.

While the evolution of our model was constrained in such a way that it could not implement it, there is another solution for filtering out interference of a sufficiently high timescale relative to the frequency of environmental stimuli that peak interference is guaranteed to coincide with all instances of environmental stimuli.
The maximum bias of nodes in our model was constrained to the maximum weight of a single incoming connection (5), which is lower than the product of the environmental intensity factor with the input scaling factor applied to inputs to the sensor neurons ($5 \times 5 = 25$).
However, a sufficiently high bias (around 12) can indeed induce the sensor neurons' output function to only be maximised when environmental stimulation is high.

These two ways of adjusting the neural biases demonstrate how a large difference in timescale between environmental signal and interference means that over time it is easy to extract the environmental signal from the summation of the two.
The activity of this system actually emphasise this difference in timescale, as typical motor activity is constrained to higher absolute ranges than the ancestral solution - see Figure \ref{fig_motor_raincloud}.
Due to the way this time dependent interference periodically saturates the input neurons, the system is not sensitive to environmental stimuli spikes that are of sufficiently low duration to perfectly coincide with motor interference peaks as the corresponding input neuron's output function would already be saturated.
Spikes of this duration do reliably induce a motor response in the other systems we've examined in this paper.
This represents a problem for the ancestral solution's strategy of taking advantage of sharp spikes in the right sensor.
Significantly, this system's Type 1 orbit is much slower than the ancestor's, with the periods of environmental stimulation of the sensor lasting for longer.
When it comes to distinguishing environmental and self-caused stimuli, the motor activity of the system not only shapes the self-caused stimuli to facilitate this, it shapes the environmental stimuli too.

As with the unavoidable squared interference, the behaviour of this system depends on the presence of its motor-driven interference.
For example, with the left motor-sensor interference removed, environmental stimulation of the left sensor inhibits rather than excites the activation of both motors. Significantly, in the absence of environmental stimulation the motor activity and corresponding interference of this system features a long transient before settling into lower magnitude oscillations, and this transient is restarted by environmental stimuli.
This effect can be seen in Figure \ref{fig_sinusoidal_late_spike}.
These prolonged effects of momentary environmental stimulation are not a feature of the previously examined systems.
They mean that the frequency of the motor driven interference varies significantly both during the approach to the light and during Type 1 orbits.
Altogether these qualities demonstrate that the evolved behaviour of this system depends on its motor-driven interference, emphasising that even interference as seemingly unruly as this can be incorporated into successful behaviour.
emphasising that even this constantly varying, time dependent interference can be incorporated into the system's behaviour.

To summarise, this system has the ability to respond to environmental stimuli despite continually varying sinusoidal interference.
Because of the difference in timescale between the frequency of the sinusoidal interference and the frequency of environmental stimulation,
a CTRNN neuron can be parametrised such that the maximisation of its output function only coincides with environmental stimuli, or such that the minimisation of its output function only coincides with the absence of environmental stimuli.
Rather than predicting and subtracting self-caused stimuli on the timescale of actions, this is a fixed solution that is implemented at the evolutionary timescale.
Furthermore, rather than subtracting out the motor-driven interference, the behaviour of the system is deeply entangled with it, displaying oscillatory motor activity driven by the interference and prolonged transient motor activity following activation of the motors in response to stimuli.
Additionally, whether an environmental stimulus is excitatory or inhibitory depends respectively on the presence or absence of motor-driven sensor stimulation.
This demonstrates that rather than suppressing self-caused stimuli, proper functioning for some systems relies on the presence of self-caused stimuli.

\section{Discussion}
\label{sec_discussion}

The canonical explanation of the sensory attenuation effect is that self-caused sensory stimuli are predicted internally using a copy of the relevant neural outputs, and then subtracted out of the sensory inputs.
This may well be the case, but our results suggest that there are other possibilities.
We have shown that a neural network controller can be successfully adapted to handle several different forms of motor-driven sensory interference.
Significantly, the adaptations we have catalogued here do not rely on \textit{predicting} this interference.
We now summarise these adaptations.

\textbf{Avoidance:} When self-caused sensory interference is only triggered by certain motor outputs,
and if the task at hand can be accomplished while avoiding those outputs,
it may be easiest for a control system to simply modify its behaviour to avoid motor-sensor interference.
We see this emerge when our model was evolved with sigmoidal interference.
In the special case where there are multiple independent sensors and motors, where each motor interferes with only one sensor, an alternative solution is possible.
If the task can be accomplished using only one sensor, then only one source of interference needs to be regulated.
Doing so permits the other motors to operate freely over a wider range of activity.
We describe this strategy as `keeping one eye on the prize'.
This is arguably just avoiding the interference, with extra steps. 
We again see this strategy used in the case of sigmoidal interference.

Where interference is unavoidable but the magnitude of the interference does depend on motor activity, motor activity can be constrained to ranges that limit the quantity of interference, reducing its magnitude relative to environmental stimuli.
This is used in the case of the unavoidable squared interference.

Minimisation and avoidance could be seen as special cases of causing the interference to plateau at a constant value. If interference is additive and non-saturating, as it is in our model, it can be eliminated by simply subtracting a constant term from the input. In general this is trivial for a CTRNN. However even without subtracting the interference out directly, constant interference just shifts the environmental stimuli's contribution to the sensor to a higher range, which does not actually change the information available when the interference is non-saturating.

\textbf{Coordination:} The timing of motor-driven interference with a sensor may be regulated to coincide with environmental stimulation of that same sensor.
One way to look at this is that the detection of a sufficiently `loud' environmental stimuli renders any coincident interference irrelevant.
With a one dimensional sensor like those used in this model, the interference is actually constructive. The coincidence of motor-driven and environmental stimuli amplifies the effect of the environmental stimuli on the sensor.
If the response to this stimuli tends to diminish the stimuli (negative feedback),
as we see when stimulation of the sensor sensor causes the robot to turn away from the light,
then this strategy of coordination can play a powerful role in establishing a stable relationship with environmental stimuli.
This can be effectively combined with a strategy of avoiding or minimising interference,
which we see with the squared interference function.

\textbf{Time scale differences:} The previous solutions don't work for interference which is continually varying in such a manner that the interference's minima and maxima are not not under direct control of the motors.
However if such interference is of a high enough frequency relative to the frequency of environmental stimuli, then this difference in time scale can be leveraged to separate the interference from the environmental stimuli.
Slowly varying stimuli can be perceived through quickly varying interference,
which we see with the sinusoidal interference function.
The evolved behaviour we saw with this interference function elevated the frequency of motor-driven stimulation further, amplifying this differential.

\textbf{Shaping environmental stimuli:}
Time scale differences are a case of natural differences between the characteristics of the interference and the environmental stimuli.
So far we've described how the system can shape the interference to minimise its negative effects or make it easier to distinguish from the environmental stimuli.
However the ancestral solution demonstrates that the shape that environmental sensor stimulation takes depends on the system's activity - sharp spikes in sensor stimulation are produced by passing close to the light while driving backwards.
With the sinusoidal interference function, we found that sharp spikes could be lost in the high frequency interference,
and that in addition to the system's behaviour raising the frequency of the motor-driven interference,
its behaviour also lowered the frequency of environmental stimulation.
Embodied systems can reliably respond differently to environmentally and self-caused stimuli because the characteristics of both forms of stimuli are at least partially determined by the system's own activity.

Removing motor-driven interference from a system optimised to perform a task in the presence of that interference does not necessarily improve interference, and may instead degrade it significantly. 
Instead the successful phototactic behaviour of the systems we've studied often incorporates interference functionally.
Coordination of interference with environmental stimuli is one case of this, where the coordination amplifies the stimuli,
but we also saw how the response to environmental stimulation of one sensor can be mediated by motor-driven stimulation of the contralateral sensor.
This suggests that it is a mistake to view the problem of coping with self-caused sensory stimuli as primarily about subtracting out the interference -
even viewing it in terms of perceiving the environment clearly despite the interference may be going too far.
It's natural to think of the phototaxis task this way,
but the evolutionary algorithm we used selected purely for phototactic ability,
and as we've seen this can involve incorporating motor-driven interference into behaviour.

This all suggests that prediction and subtraction do not tell the whole story when it comes to coping with self-caused sensory stimuli.
In some ways this is obvious, as self-caused sensory stimuli are involved in a range of activities in which they do not play an interfering role.
For example, the sensation of self-touch when kneading an aching muscle, or occlusion of the visual field when engaging in visually guided reaching and grasping.
In these activities, self-caused sensory stimuli are actually desirable.
Nevertheless, our model shows that even in situations where clear perception of the environment is \textit{prima facie} desirable, self-caused sensory stimuli may not play an entirely interfering role.
Furthermore, we see that even when responsiveness to the environment is needed, prediction and subtraction are not the only game in town.

How do these results actually relate to the predictive account of coping with self-caused stimuli?
A criticism of our results may be that the problems we have posed our model are insufficiently `representation-hungry' to require prediction.
Representation-hungry problems are those that seem to require the use of internal representations to be solved, 
defined by \cite{clark_doing_1994} to be a problem where one or both of the following conditions hold.
Condition one is that the problem involves reasoning about \textit{absent}, non-existent, or counterfactual states of affairs.
Condition two is that the problem demands selective sensitivity to parameters whose sensory manifestations are `complex and unruly' - 
that is, the system must be able to treat differently inputs whose sensory manifestations are highly similar, and conversely be able to treat similarly inputs whose sensory manifestations are very different.
We actually agree that our model does not solve a representation-hungry problem, and in fact see this is a primary contribution of our results.
In general, coping with self-caused sensory stimuli need not be a representation-hungry problem.

How we process self-caused stimuli is typically taken to involve an internal predictive model.
Prediction itself is a task which meets the first condition, since prediction inherently involves states of affairs that do not yet exist.
However the fundamental problem the predictive model is being used to solve meets only the second criteria, that is treating differently self-caused and externally-caused inputs whose sensory manifestations may be identical.
Otherwise identical inputs can be distinguished by predicting, based on an internal model, whether an input is self-caused or externally-caused.
If prediction is necessary then the problem of coping with self-caused stimuli would seem to meet the criteria for representation-hunger.

The way self-caused stimuli have been studied experimentally highlights what we see as a key limitation of the representational paradigm.
In the lab, where the stimuli presented to the subject are fully controlled by the experimenter and the subject's activities are highly constrained,
we are forced to take sensory stimuli as given.
This essentially enforces a condition where the outcome of the experiment is determined by mental activity decoupled from meaningful and ongoing sensorimotor feedback.
Under such circumstances, a strict interpretation of Clark and Toribio's second criteria may hold - where self-caused and externally-caused stimuli are identical to the extent that only knowledge over and above their sensory manifestations can distinguish them.
However the everyday problem of coping with self-caused sensory stimuli occurs outside the lab,
where these stimuli are part of our ongoing sensorimotor activity.
In this case our model has shown that there are diverse ways to perform successfully and even to distinguish self-caused and externally-caused stimuli.
A key part of this is that both types of stimuli are shaped by our own activity, and thus encountered on our own terms.
In these circumstances, the strict definition is unlikely to hold, as we can shape both self and externally caused stimuli to differentiate them.

While the problem of distinguishing truly identical sensory inputs may well be representation-hungry, our model's embodiment allows it to shape its inputs such that they are distinguishable by non-predictive means.
Thus we grant that our model does not capture a strictly representation-hungry problem, a conclusion directly supported by our results.
This is a not a limitation of this study, it's a feature.
Our model shows that representational cognition is not necessary in general to cope with self-caused stimuli, because of the capabilities afforded by embodiment.
In effect, this shrinks the set of human capabilities which are taken to require representational cognition.

The idea of representation-hunger highlights a long running critique of embodied cognition,
where solving tasks in representation-free, embodied ways aren't considered central examples of what we really mean by cognition.
A distinction is drawn between tasks solvable via online and potentially representation-free sensorimotor processing,
and offline cognition operating on internal, representational models \citep{zahnoun_on_2019}.
It is worth noting that similarly minimal, CTRNN controlled models have successfully solved problems with requirements like memory without the use of internal representations. \cite{beer_information_2015} demonstrate how a robot can both remember a cue and categorise a subsequent probe relative to that cue by offloading memory to the environment and structuring its relationship with its environment to facilitate direct perception on the relative difference between cue and probe.
It was only when the robot's ability to move while being presented with the cue was removed that information about the cue was retained internally in the neural activation.
Studies like this push back at the idea that internal representation is necessary to solve problems requiring responses to abstract or absent stimuli, by showing that other possibilities are facilitated by the way embodiment structures the ongoing relationship between controller and environment.

\section*{Conflict of Interest Statement}

The authors declare that the research was conducted in the absence of any commercial or financial relationships that could be construed as a potential conflict of interest.

\section*{Author Contributions}

J.G. and M.E. conceived of the idea for this project together. J.G wrote the code,
ran the experiments, analysed the data, produced the figures and wrote the manuscript. M.E. provided feedback on the experiments and on multiple drafts of the manuscript.

\section*{Acknowledgments}

The authors wish to acknowledge the use of New Zealand eScience Infrastructure (NeSI) high performance computing facilities, consulting support and/or training services as part of this research. New Zealand's national facilities are provided by NeSI and funded jointly by NeSI's collaborator institutions and through the Ministry of Business, Innovation \& Employment's Research Infrastructure programme. URL https://www.nesi.org.nz.

Figures were produced with Matplotlib \citep{hunter_matplotlib_2007}. The idea for the `raincloud' presentation of the data used in Figure \ref{fig_motor_raincloud} is due to \cite{allen_raincloud_2021}.

An extended abstract summarising parts of this work has been submitted to the ALIFE 2022 conference.

\bibliographystyle{apalike}

\end{document}